%% file: main.tex
\documentclass{article}

\PassOptionsToPackage{numbers, sort&compress}{natbib}

\usepackage[final]{neurips_2025}

\input{preamble}

\AtEndPreamble{
    \usepackage[capitalize]{cleveref}
    \crefname{section}{Sec.}{Secs.}
    \Crefname{section}{Section}{Sections}
    \Crefname{table}{Table}{Tables}
    \crefname{table}{Tab.}{Tabs.}
}

\usepackage[utf8]{inputenc} 
\usepackage[T1]{fontenc}    
\usepackage{hyperref}       
\usepackage{url}            
\usepackage{booktabs}       
\usepackage{amsfonts}       
\usepackage{nicefrac}       
\usepackage{microtype}      
\usepackage{xcolor}         

\title{FlowFeat: Pixel-Dense Embedding of Motion Profiles}

\author{%
   Nikita Araslanov\textsuperscript{\,1 2}
   \And
   Anna Sonnweber\textsuperscript{\,1}
   \And
   Daniel Cremers\textsuperscript{\,1 2} \\
   \AND
   \normalfont\textsuperscript{1}TU Munich \qquad \textsuperscript{2}MCML
}

\begin{document}

\maketitle

\vspace{-2em}
\begin{figure}[h]
\captionsetup[subfigure]{labelformat=empty}
\centering

\begin{subfigure}{0.161\linewidth}
\caption{\footnotesize Input}
\includegraphics[width=\linewidth]{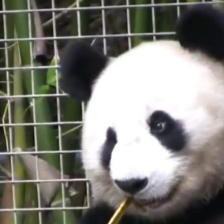}
\end{subfigure}
\hfill
\begin{subfigure}{0.161\linewidth}
\caption{\footnotesize  DINOv2 \cite{Oquab:2023:DINOv2}}
\includegraphics[width=\linewidth]{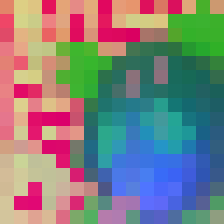}
\end{subfigure}
\hfill
\begin{subfigure}{0.161\linewidth}
\caption{\footnotesize  V-JEPA \cite{Bardes:2024:RFP}}
\includegraphics[width=\linewidth]{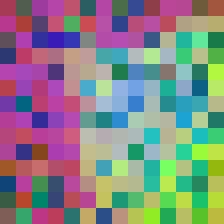}
\end{subfigure}
\hfill
\begin{subfigure}{0.161\linewidth}
\caption{\footnotesize  VideoMAE \cite{tong:2022:videomae}}
\includegraphics[width=\linewidth]{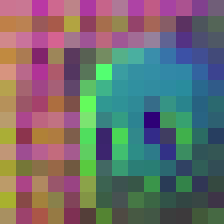}
\end{subfigure}
\hfill
\begin{subfigure}{0.161\linewidth}
\caption{\footnotesize  FeatUp \cite{Fu:2024:FeatUp}}
\includegraphics[width=\linewidth]{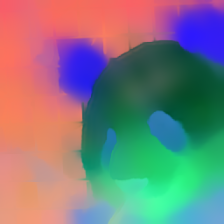}
\end{subfigure}
\hfill
\begin{subfigure}{0.161\linewidth}
\caption{\textbf{\footnotesize  \oursName (ours)}}
\includegraphics[width=\linewidth]{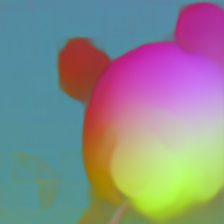}
\end{subfigure}

\begin{subfigure}{0.161\linewidth}
\includegraphics[width=\linewidth]{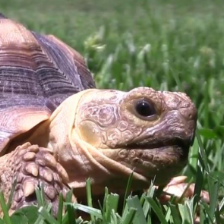}
\end{subfigure}
\hfill
\begin{subfigure}{0.161\linewidth}
\includegraphics[width=\linewidth]{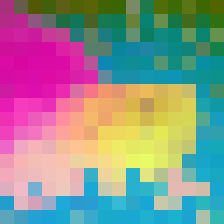}
\end{subfigure}
\hfill
\begin{subfigure}{0.161\linewidth}
\includegraphics[width=\linewidth]{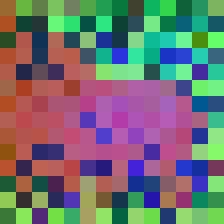}
\end{subfigure}
\hfill
\begin{subfigure}{0.161\linewidth}
\includegraphics[width=\linewidth]{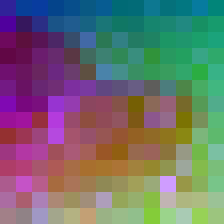}
\end{subfigure}
\hfill
\begin{subfigure}{0.161\linewidth}
\includegraphics[width=\linewidth]{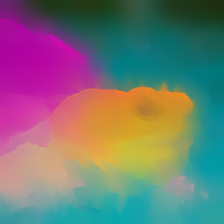}
\end{subfigure}
\hfill
\begin{subfigure}{0.161\linewidth}
\includegraphics[width=\linewidth]{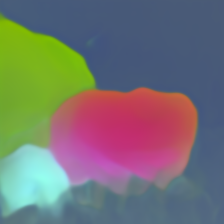}
\end{subfigure}

\begin{subfigure}{0.161\linewidth}
\includegraphics[width=\linewidth]{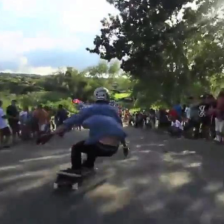}
\end{subfigure}
\hfill
\begin{subfigure}{0.161\linewidth}
\includegraphics[width=\linewidth]{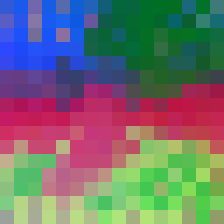}
\end{subfigure}
\hfill
\begin{subfigure}{0.161\linewidth}
\includegraphics[width=\linewidth]{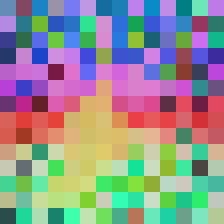}
\end{subfigure}
\hfill
\begin{subfigure}{0.161\linewidth}
\includegraphics[width=\linewidth]{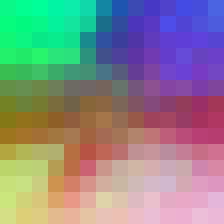}
\end{subfigure}
\hfill
\begin{subfigure}{0.161\linewidth}
\includegraphics[width=\linewidth]{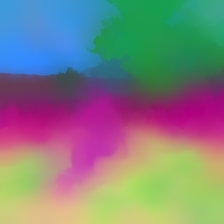}
\end{subfigure}
\hfill
\begin{subfigure}{0.161\linewidth}
\includegraphics[width=\linewidth]{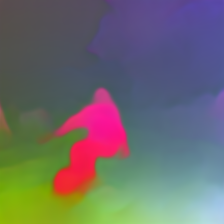}
\end{subfigure}

\caption{\textbf{\oursName} is a versatile feature representation at pixel-level resolution. Embedding profiles of plausible motion, \oursName stands out from existing techniques by offering excellent spatial precision coupled with temporal consistency. Here, we visualise (using PCA with three principal components) a comparison of \oursName with the feature maps of the state-of-the-art vision encoders.}
\label{fig:flow_approx}
\vspace{-0.5em}
\end{figure}

\begin{abstract}
Dense and versatile image representations underpin the success of virtually all computer vision applications. However, state-of-the-art networks, such as transformers, produce low-resolution feature grids, which are suboptimal for dense prediction tasks. To address this limitation, we present \oursName, a high-resolution and multi-task feature representation. The key ingredient behind \oursName is a novel distillation technique that embeds a distribution of plausible apparent motions, or \emph{motion profiles}. By leveraging optical flow networks and diverse video data, we develop an effective self-supervised training framework that statistically approximates the apparent motion. With its remarkable level of spatial detail, \oursName encodes a compelling degree of geometric and semantic cues while exhibiting high temporal consistency. Empirically, \oursName significantly enhances the representational power of five state-of-the-art encoders and alternative upsampling strategies across three dense tasks: video object segmentation, monocular depth estimation and semantic segmentation.
Training \oursName is computationally inexpensive and robust to inaccurate flow estimation, remaining highly effective even when using unsupervised flow networks.
Our work takes a step forward towards reliable and versatile dense image representations.\blfootnote{Project website: \href{https://tum-vision.github.io/flowfeat}{https://tum-vision.github.io/flowfeat}.}\blfootnote{Code and pre-trained models (Apache-2.0 License): \href{https://github.com/tum-vision/flowfeat}{https://github.com/tum-vision/flowfeat}.}
\end{abstract}

\input{sections/01_intro}
\input{sections/02_related}
\input{sections/03_method}
\input{sections/04_experiments}
\input{sections/05_conclusion}

{
    \small
    \bibliographystyle{abbrvnat}
    \bibliography{main}
}


\clearpage
\pagenumbering{roman}
\appendix

\input{supp-arxiv}

\section{License note}
\label{sec:license}
The parts of the code we use from \citet{Jabri:2020:STC} (the label propagation algorithm) are released under a MIT license.
The datasets YouTube-VOS, Kinetics-400 are licensed under the Creative Commons Attribution 4.0 International License, while DAVIS-2017 is provided under the Creative Commons Attribution-NonCommercial 4.0 International License.

\end{document}

%% file: preamble.tex
\usepackage{graphicx}
\usepackage{bbding}
\usepackage{tabularx}
\usepackage{booktabs}
\usepackage{multirow}
\usepackage{subcaption}
\usepackage{microtype}

\usepackage{wrapfig}

\usepackage{etoolbox,siunitx}
\robustify\bfseries
\usepackage{listings}
\usepackage{xcolor}
\usepackage{colortbl}

\definecolor{codegreen}{rgb}{0,0.6,0}
\definecolor{codegray}{rgb}{0.5,0.5,0.5}
\definecolor{codepurple}{rgb}{0.58,0,0.82}
\definecolor{backcolour}{rgb}{0.95,0.95,0.92}
\definecolor{azure}{rgb}{0.94, 1.0, 1.0}
\definecolor{lightgrey}{rgb}{0.6, 0.6, 0.6}

\lstdefinestyle{mystyle}{
    backgroundcolor=\color{backcolour},
    commentstyle=\color{codegreen},
    keywordstyle=\color{magenta},
    numberstyle=\tiny\color{codegray},
    stringstyle=\color{codepurple},
    basicstyle=\ttfamily\scriptsize,
    breakatwhitespace=false,
    breaklines=true,
    captionpos=b,
    keepspaces=true,
    numbers=left,
    numbersep=5pt,
    showspaces=false,
    showstringspaces=false,
    showtabs=false,
    tabsize=2
}

\usepackage{pifont} 
\RequirePackage{xspace}
\makeatletter
\DeclareRobustCommand\onedot{\futurelet\@let@token\@onedot}
\def\@onedot{\ifx\@let@token.\else.\null\fi\xspace}

\def\eg{\emph{e.g}\onedot} 

\def\ie{\emph{i.e}\onedot} 

\def\cf{\emph{cf}\onedot}

\def\wrt{w.r.t\onedot}

\makeatother

\newcommand*{\inparagraph}[1]{\noindent\textbf{#1}\hspace{0.5em}}

\newcommand{\oursName}{FlowFeat\@\xspace}
\newcommand{\oursNamePP}{FlowFeat++\@\xspace}
\newcommand{\oursNameX}[1]{FlowFeat-{#1}\@\xspace}
\newcommand{\oursNameXPP}[1]{FlowFeat-{#1}++\@\xspace}
\newcommand{\supp}{supp.\ material\@\xspace}
\newcommand{\leftPad}{\: +\,}
\newcommand{\JF}{$\mathcal{JF}$\@\xspace}

\definecolor{plusgreen}{rgb}{0,0.6,0}
\definecolor{minusred}{rgb}{0.9,0.0,0.0}
\newcommand{\minus}[1]{\color{minusred}-#1}
\newcommand{\plus}[1]{\color{plusgreen}{+#1}}

\newcommand\blfootnote[1]{%
  \begingroup
  \renewcommand\thefootnote{}\footnote{#1}%
  \addtocounter{footnote}{-1}%
  \endgroup
}

%% file: sections/01_intro.tex
\section{Introduction}
\label{sec:intro}

The feature maps of state-of-the-art self-supervised encoders (\eg \cite{He:2022:MAA,Caron:2021:EPS}) have drastically downsampled spatial resolutions (\eg by a factor of $16$), as illustrated in \cref{fig:flow_approx}.
While such downsampling improves the computational efficiency of deep networks, it compromises on the accuracy of dense prediction tasks, where spatial detail is crucial.
Upsampling techniques, such as those based on bilateral filters \cite{Fu:2024:FeatUp}, can recover feature detail to an impressive degree.
However, bilateral upsampling incurs a tangible computational cost and struggles under challenging illumination scenarios (\cf \cref{fig:flow_approx}, third row).
Alternatively, one could equip encoders with a lightweight decoder module, such as DPT \cite{Ranftl:2021:VTD}.
However, training such decoders without human annotation is highly non-trivial.
Building on this motivation, we present \emph{\oursName}, a multi-task pixel-level image representation obtained in a label-efficient (or even label-free) manner.

Different from much of the existing work on representation learning, \oursName derives from dense motion patterns rather than the static appearance alone \cite{He:2022:MAA, Oquab:2023:DINOv2,CaronHS24,Assran:2023:SSL}. While \oursName is a monocular model operating on a \emph{single} input image at test time, it uses unlabelled videos for training to embed motion patterns into a pixel-level representation.
Motion patterns are foundational to visual perception \cite{marr1982representation}; they encode the compositional nature of visual scenes, encompassing both semantically and geometrically meaningful phenomena.
However, as \cref{fig:flow_approx} illustrates, video-based learning still fails to provide representations that are dense, versatile and effective \cite{Jabri:2020:STC,Feichtenhofer:2022:MAA,Bardes:2024:RFP}.

As a step forward, we synergise state-of-the-art optical flow networks and real-world video data.
On the one hand, modern optical flow networks produce dense motion estimates with outstanding accuracy, even in challenging settings \cite{TeedD20,Wang:2024:SEA,Sun:2018:PWC}.
On the other hand, datasets of casual videos provide a treasure trove of motion and scene diversity \cite{Xu:2018:YTV,Kay:2017:KHA}.
Combining both ingredients in a joint learning framework, \oursName requires no manual annotation.
Optical flow networks train predominantly on synthetically generated labels or even with self-supervision \cite{Meister:2018:UFU,Stone:2021:SMURF}; video datasets derive from real-world benchmarks and require minimal curation (\eg montage filtering).
The key technical challenge is distilling the apparent motion in a fashion accommodating its stochastic nature.

\oursName addresses this challenge with a simple idea.
We estimate the feature representation with a \emph{distribution} of linear transformations.
Intuitively, for a given image and a flow estimate \wrt a randomly sampled counterpart, \oursName is trained to admit a linear transformation approximating the flow.
Specifically, every training iteration estimates a \emph{lower bound} of this transformation \emph{on-the-fly} using a least-squares formulation.
The statistical nature of this lower-bound approximation (due to sampling of the image pair) accommodates motion stochasticity and proves crucial for dealing with inaccurate flow and occasional static scenes.
Consequently, the distribution of linear transformations allows \oursName to embed a distribution of plausible motion, or \emph{motion profiles} \cite{Shi:1998:MST}.

Overall, our work presents two contributions.
First, we develop an effective self-supervised training framework that exploits the synergetic power of flow networks and large video datasets to embed motion profiles.
Our framework is efficient at training time and can run comfortably within academic infrastructures.
Second, we extensively evaluate the learned representation, \oursName, on three diverse tasks of dense prediction: video object segmentation (VOS), monocular depth estimation and semantic segmentation.
Our analyses reveal a consistent benefit of \oursName across all tasks, exhibiting a compelling degree of temporal consistency and spatial detail.
Furthermore, \oursName has appealing practical properties: \textit{(i)} it is runtime- and label-efficient; \textit{(ii)} it scales well with varying input resolution without the need for model fine-tuning, and  \textit{(iii)} it facilitates simple post-processing tasks, enhancing the quality of dense predictions without additional training.

%% file: sections/02_related.tex
\section{Related Work}
\label{sec:related}

A substantial effort towards unsupervised feature representations has focused on learning from large image sets \cite{Chen:2020:ASF,GrillSATRBDPGAP20,Assran:2023:SSL}.
This development spans multiple axes of pursuit, such as model efficiency \cite{Caron:2021:EPS,ZbontarJMLD21}, scalability \cite{He:2022:MAA,Oquab:2023:DINOv2} and framework architecture \cite{Chen:2021:ESS}.
Although pre-training from image sets dominates the research landscape in unsupervised learning, there have been natural extensions of image-based frameworks to learning from video data \cite{Feichtenhofer:2022:MAA,Bardes:2024:RFP,tong:2022:videomae}.
However, it remains challenging to obtain \emph{spatio-temporal} representations that are both dense (\ie pixel-level) and temporally consistent \cite{Wang:2019:LCC,Araslanov:2021:DUL,CaronHS24}.
Central to learning spatio-temporal representations is the design of the pre-text task. 
One prominent technique is \emph{cycle consistency} \cite{Wang:2019:LCC,Lai:2019:SSV,Jabri:2020:STC,Son22}.
It constructs a temporal palindrome from a video sequence, ensuring consistency of a putative state in a forward and backward directions.
Contrastive learning underpins another broad category of the research effort \cite{Pinheiro:2020:ULD}.
The main ideas are: constructing a reliable set of positive and negative samples \cite{Jeon:2021:MBS}; combining learning on pixel, frame, and video levels \cite{Wang:2021:CTS,Xu:2021:RSC}; or jointly representing a video clip with a limited set of contrastive anchors \cite{Araslanov:2021:DUL}.
Unlike these feature-based techniques, which have limited resolution, photometric-based learning, such as colourisation, relies on natural radiance-based appearance \cite{VondrickSFGM18}.
\citet{Lai:2020:MAS} leverage this technique in video-based learning, reconstructing the target frame from previous frames observed in the CIELAB colour space.

Feature upsampling strategies, such as FeatUp \cite{Fu:2024:FeatUp} and LoftUp \cite{Haiwen:2025:LUP} are closely related to our work.
In contrast to bilateral upsampling \cite{Kopf:2007:JBU,Fu:2024:FeatUp}, \oursName is more computationally efficient and has \emph{complementary} properties to the low-resolution encoder features.
Unlike contemporaneous work \cite{Haiwen:2025:LUP} leveraging SAM \cite{kirillov:2023:SAM}, \oursName is label-efficient and can be trained in an unsupervised manner.

Representation learning by or with motion estimation is not new \cite{Pathak:2017:LFW,Mahendran:2018:CPO,Han:2020:SSC} and traces back to the earlier works on trajectory clustering and motion-based segmentation \cite{ZM:2003:DDI,Liu:2005:MMa,Brox:2010:OSL,Fragkiadaki:2015:LSM}.
Training \oursName is efficient, since it does not require pairwise sampling \cite{Mahendran:2018:CPO}; nor does it require object discovery \cite{Pathak:2017:LFW,Henaff:2022:ODIN}.
Instead, \oursName learns directly from optical flow provided from off-the-shelf networks with a distribution of linear transformations.
This approach takes primary inspiration from motion profiles, which model a distribution of velocities at a given pixel \cite{Shi:1998:MST}.
\emph{Embedding} motion profiles, \oursName enhances downstream accuracy of the baseline representation across diverse tasks.

%% file: sections/03_method.tex
\begin{figure*}[t]
    \centering
    \begin{minipage}{.6\textwidth}
    \def\svgwidth{1.0\linewidth}
    \input{figures/framework.tex}
    \end{minipage}%
    \hfill
    \begin{minipage}{.35\textwidth}
\begin{lstlisting}[language=Python,mathescape=true]
# compute optical flow
F = FlowNet(frame, frame_other)
# random crop transforms
tf1, tf2 = random_crops()
# two views of the first frame
v1 = crop(frame, tf1)
v2 = crop(frame, tf2)
u1,u2 = crop(F,tf1),crop(F,tf2)
# dense feature maps
x1 = dec_ema(encoder(v1))
x2 = dec(encoder(v2))
# Eq. 4: teacher optimal A
A$^\ast$ = lstsq(x1, u1)
# student predicted flow
u2$^\ast$ = x2 @ A$^\ast$
# Eq. 7: flow loss w.r.t. dec
loss = flow_loss(u2$^\ast$, u2)
loss.backward()
# EMA update
dec_ema.update_from(dec)
\end{lstlisting}
    \end{minipage}
\caption{\textbf{Embedding motion profiles}: \oursName relies on the exponentially moving average (EMA) teacher model and learns to reconstruct apparent motion with a distribution of linear transformations. For a given frame $I_t$, we randomly sample its temporal counterpart $I_{t^\prime}$. A pre-trained network $\mathcal{F}$ computes optical flow $F_{(t \rightarrow t^\prime)}$. We generate two overlapping random crops of frame $I_t$ and feed the resulting views $v_1$ and $v_2$ to the teacher and the student networks, respectively. Obtaining the optimal linear transform $A^\ast$ on-the-fly with ridge regression in the teacher branch, we compute the reconstruction loss \wrt the flow crop $u_2$ to update the student parameters $\theta$ with gradient descent.}
\label{fig:framework}
\vspace{-0.8em}
\end{figure*}
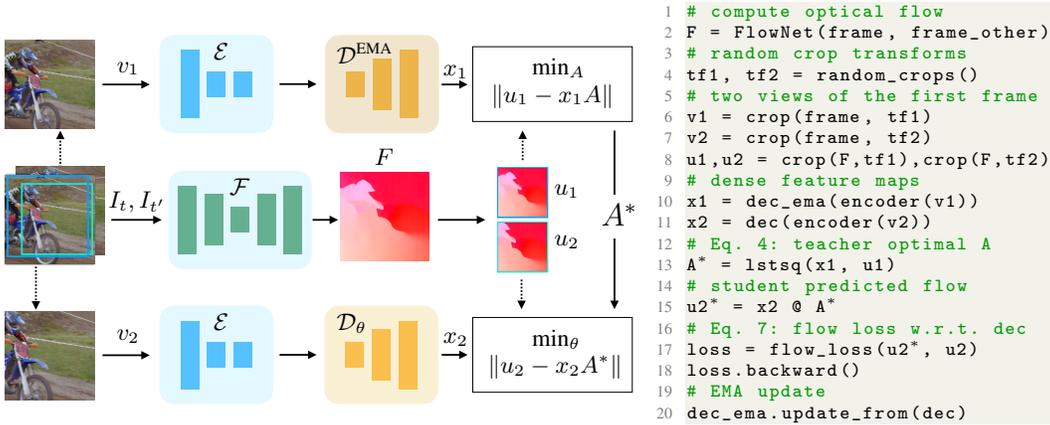

\section{Embedding Motion Profiles}
\label{sec:method}

\inparagraph{Linear maps for optical flow.} To obtain pixel-level features enhancing the low-resolution representation of pre-trained encoders, we estimate apparent motion in real-world video sequences.
Off-the-shelf optical flow models exhibit exceptional generalisation, despite being trained on synthetic scenes \cite{TeedD20,Wang:2024:SEA} or even with self-supervision \cite{Stone:2021:SMURF}.
However, distilling motion estimates into a \emph{monocular} model (in contrast to previous work \cite{Liu:2022:LDA}), is highly non-trivial due to motion stochasticity.\footnote{Na\"{i}vely approximating optical flow with a single linear layer unsurprisingly fails, as we verify in \cref{sec:ablation}.}
Overcoming this issue, we train an image representation $\mathcal{H}_\theta(I) = x$ such that for \emph{any} temporally neighbouring frame of $I$, there exists a linear operator on $x$ which approximates the optical flow \wrt that neighbour.
Since we estimate the linear operator uniquely for each frame neighbour, the learned representation $x$ would embed \emph{statistical motion patterns} for each input image $I$ -- an idea inspired by motion profiles \cite{Shi:1998:MST}.

Given image $I_t$ and its temporal neighbour $I_{t^\prime}$ of resolution $H \times W (=: N)$, we formulate the idea above with the following flow reconstruction objective (where $\|\cdot \|$ denotes an ``entry-wise'' norm):
\begin{equation}\label{eq:problem}
    \text{min}_{\theta,A} \quad \mathbb{E}_{I_t,I_{t^\prime}} \big[ \|\mathcal{F}(I_t, I_{t^\prime}) - \mathcal{H}_\theta(I_t) A \| \big],
\end{equation}
where $\mathcal{F}(I_t,I_{t^\prime}) \in \mathbb{R}^{N\times 2}$ is the optical flow from a pre-trained network \cite{TeedD20,Wang:2024:SEA}; $\mathcal{H}_\theta(I_t) \in \mathbb{R}^{N \times d}$ is our learned pixel-level feature representation and $A \in \mathbb{R}^{d \times 2}$ is a linear operator.
Note that since $\mathcal{H}_\theta$ and $A$ are both unknown, \cref{eq:problem} is an ill-posed problem due to scale ambiguity.\footnote{If $A^\ast$ and $\mathcal{H}^\ast$ are the solutions, so are $c A^\ast$ and $\mathcal{H}^\ast / c$ for any $c \neq 0$.}
Therefore, we propose to compute the corresponding loss in two steps: \textit{(i)} computing a lower-bound $A^\ast$ with a surrogate teacher network, while keeping $\mathcal{H}$ fixed; \textit{(ii)} computing the gradient \wrt $\theta$ of the original network by swapping $A^\ast$ into \cref{eq:problem} as the lower-bound linear approximation.

\inparagraph{Student-teacher framework.} \cref{fig:framework} illustrates the framework and the corresponding training algorithm.
Leveraging the mean teacher as the training model \cite{Tarvainen:2017:MTB}, our network $\mathcal{H}_\theta := \mathcal{D}_\theta \circ \mathcal{E}$ comprises a fixed (pre-trained) encoder $\mathcal{E}$ and a trained lightweight decoder $\mathcal{D}_\theta$, which outputs a dense feature representation of dimensionality $d$.
The teacher model $\mathcal{H}^\text{EMA}$ is equivalent to $\mathcal{H}_\theta$ with the exception of the decoder $\mathcal{D}^\text{EMA}$, which is an exponential moving average of $\mathcal{D}_\theta$.

To construct the training batch, we sample two frames, $I_t$ and $I_{t^\prime}$, where $I_{t^\prime}$ could be selected from a temporal window around $I_t$. We first compute optical flow $\mathcal{F}(I_t, I_{t^\prime}) \in \mathbb{R}^{N \times 2}$ with a network pre-trained on synthetic data \cite{TeedD20,Wang:2024:SEA} or with self-supervision \cite{Stone:2021:SMURF}.
Generating two overlapping random crops of the first frame $I_t$, we feed the corresponding views $v_1$ and $v_2$ to the teacher and student models, respectively.
Using the teacher output, we solve a least-squares problem:
\begin{equation}\label{eq:lstsq}
    A^\ast = \text{argmin}_A \: \| u_1 - \mathcal{H}^\text{EMA}(v_1) A \|_2,
\end{equation}
where $u_1$ is the crop of the optical flow corresponding to view $v_1$.
In practice, we solve \cref{eq:lstsq} with ridge regression, which yields stable solutions in the presence of inaccurate flow estimates and improves training stability (\cf \cref{sec:ablation} for empirical results).
Specifically, we solve
\begin{equation}\label{eq:lstsq_ridge}
    \text{min}_A \: \| u_1 - \mathcal{H}^\text{EMA}(v_1) A \|_2 + \gamma \| A \|_2,
\end{equation}
in each training iteration.
Here, $\gamma$ is a ridge hyperparameter fixed for all models.
Setting $x_1 := \mathcal{H}^\text{EMA}(v_1)$ to simplify the notation, the closed-form solution of \cref{eq:lstsq_ridge} is naturally
\begin{equation}\label{eq:ridge}
    A^\ast = (x_1^T x_1 + \gamma I)^{-1} x_1^T u_1.
\end{equation}
Note that the first term has the \emph{feature} dimensions, $d \times d$, fixed to $d = 128$ in our experiments. 
Therefore, computing \cref{eq:ridge} has a negligible computational cost. 
In contrast to previous work \cite{Mahendran:2018:CPO}, our framework remains computationally efficient regardless of the image resolution.

Fixing $A^\ast$, we now formulate the flow reconstruction loss \wrt the student parameters of $\mathcal{H}_\theta$ as
\begin{equation}\label{eq:loss_flow}
    \mathcal{L}_{L1}(u_2, v_2) = \| u_2 - \mathcal{H}_\theta(v_2) A^\ast \|_1.
\end{equation}
The loss encourages the two overlapping crops of an input frame to admit the \emph{same} linear mapping \(A^\ast\) from the features to optical flow, thereby promoting grouping of pixels with similar motion patterns.
Note that for zero motion (\ie static scenes) the solution is \(A^\ast = 0\), which yields zero gradient for the reconstruction term, effectively discarding such training samples in the learning process.
As we also verify in the ablation study (\cf \cref{tab:ablation}), ridge regularization and the robust $L_1$ loss improve resilience of the framework to inaccuracies in the estimated target flow $u_1$ and $u_2$, respectively.

\inparagraph{Focal gradient matching.}
Motion boundaries in optical flow are well-known to reveal semantic and geometric scene components.
Therefore, we promote flow consistency at motion boundaries with an auxiliary second-order term implementing \emph{focal} gradient matching:
\begin{equation}\label{eq:edge}
    \mathcal{L}^x_{\nabla}(u_2, u^\ast_2) = (1 - e^{-\nabla_x u_2 / \sigma}) \| \nabla_x u_2 - \nabla_x u^\ast_2 \|_1,
\end{equation}
where $u^\ast_2 := \mathcal{H}_\theta(v_2) A^\ast$ and $\nabla_x$ is the spatial gradient along the $x$-axis of the image plane.
Equivalently, we compute the gradient for the $y$-axis and the corresponding term $\mathcal{L}^y_\nabla$.

\begin{figure}[t]
\captionsetup[subfigure]{labelformat=empty}
\centering

\begin{minipage}{.49\textwidth}
\begin{subfigure}{0.326\linewidth}
\caption{Input}
\includegraphics[width=\linewidth]{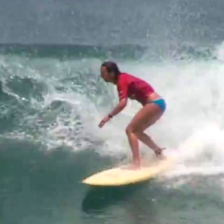}
\end{subfigure}
\begin{subfigure}{0.326\linewidth}
\caption{without $\mathcal{L}_\nabla$}
\includegraphics[width=\linewidth]{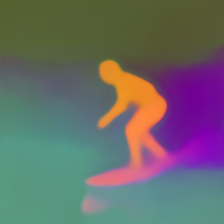}
\end{subfigure}
\begin{subfigure}{0.326\linewidth}
\caption{with $\mathcal{L}_\nabla$}
\includegraphics[width=\linewidth]{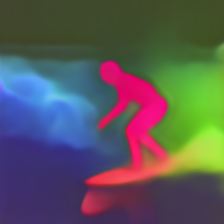}
\end{subfigure}

\vspace{0.1em}

\begin{subfigure}{0.326\linewidth}
\caption{$\sigma = 1.0$}
\end{subfigure}
\begin{subfigure}{0.326\linewidth}
\caption{$\sigma = 0.1$}
\end{subfigure}
\begin{subfigure}{0.326\linewidth}
\caption{$\sigma = 0.05$}
\end{subfigure}

\begin{subfigure}{0.326\linewidth}
\begin{picture}(0,0)
    \put(0,0){\includegraphics[width=\linewidth]{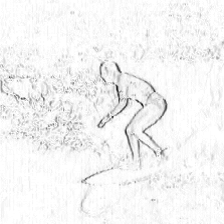}}
    \put(5,5){\scriptsize\color{black}{$\nabla_x \rightarrow$}}
\end{picture}
\end{subfigure}
\begin{subfigure}{0.326\linewidth}
\includegraphics[width=\linewidth]{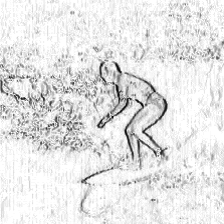}
\end{subfigure}
\begin{subfigure}{0.326\linewidth}
\includegraphics[width=\linewidth]{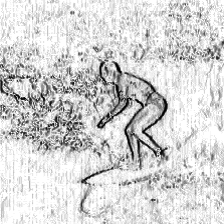}
\end{subfigure}

\begin{subfigure}{0.326\linewidth}
\begin{picture}(0,0)
\put(0,0){\includegraphics[width=\linewidth]{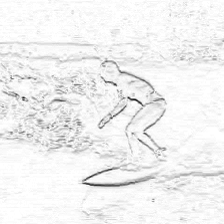}}
\put(5,5){\scriptsize\color{black}{$\nabla_y \rightarrow$}}
\end{picture}
\end{subfigure}
\begin{subfigure}{0.326\linewidth}
\includegraphics[width=\linewidth]{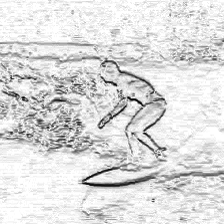}
\end{subfigure}
\begin{subfigure}{0.326\linewidth}
\includegraphics[width=\linewidth]{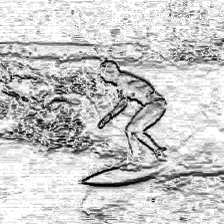}
\end{subfigure}
\end{minipage}
\hfill
\begin{minipage}{.49\textwidth}
\begin{subfigure}{0.326\linewidth}
    \includegraphics[width=\linewidth]{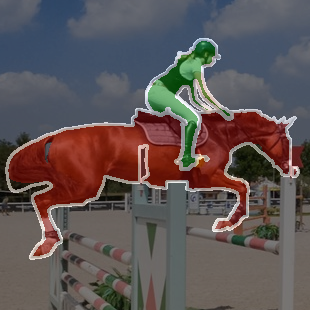}
\end{subfigure}
\begin{subfigure}{0.326\linewidth}
    \caption{Ground Truth}
    \includegraphics[width=\linewidth]{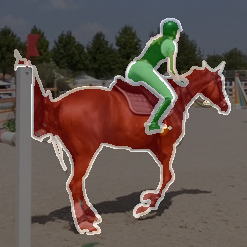}
\end{subfigure}
\begin{subfigure}{0.326\linewidth}
    \includegraphics[width=\linewidth]{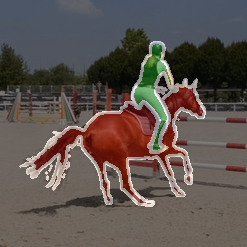}
\end{subfigure}

\begin{subfigure}{0.326\linewidth}
    \begin{picture}(64,64)
        \put(0,0){\includegraphics[width=\linewidth]{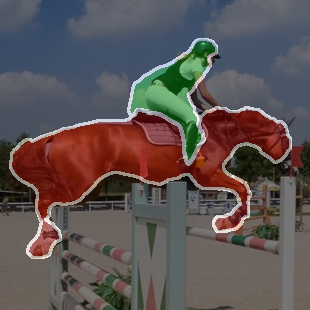}}
        \put(3,55){\scriptsize\color{white}{FeatUp $\rightarrow$}}
    \end{picture}
\end{subfigure}
\begin{subfigure}{0.326\linewidth}
    \caption{Predictions}
    \begin{picture}(64,64)
        \put(0,0){\includegraphics[width=\linewidth]{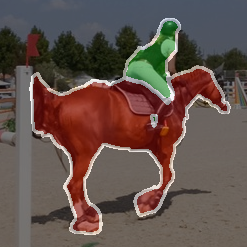}}
    \end{picture}
\end{subfigure}
\begin{subfigure}{0.326\linewidth}
    \begin{picture}(64,64)
        \put(0,0){\includegraphics[width=\linewidth]{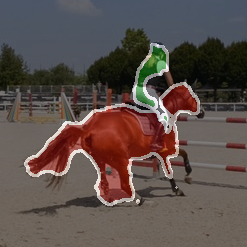}}
    \end{picture}
\end{subfigure}
\begin{subfigure}{0.326\linewidth}
    \begin{picture}(64,64)
        \put(0,0){\includegraphics[width=\linewidth]{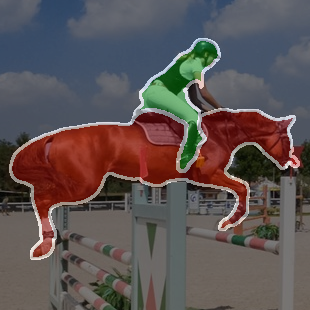}}
        \put(3,55){\scriptsize\color{white}{\oursName $\rightarrow$}}
    \end{picture}
\end{subfigure}
\begin{subfigure}{0.326\linewidth}
    \begin{picture}(64,64)
        \put(0,0){\includegraphics[width=\linewidth]{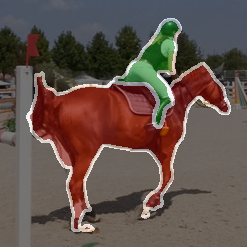}}
    \end{picture}
\end{subfigure}
\begin{subfigure}{0.326\linewidth}
    \begin{picture}(64,64)
        \put(0,0){\includegraphics[width=\linewidth]{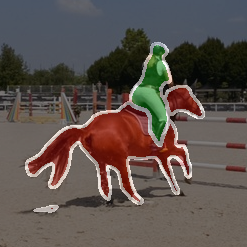}}
    \end{picture}
\end{subfigure}

\end{minipage}

\caption{\textbf{Left: Focal gradient matching term $\mathcal{L}_\nabla$.} The first row visualises the first three PCA components of \oursName trained with and without the gradient term. Observe sharper feature boundaries with the use of the gradient term.  Additionally, we found benefit in modulating the gradient difference with a hyperparameter $\sigma$, as defined in \cref{eq:edge}. The modulation with a lower $\sigma$ amplifies the effect of motion discontinuities (here, demonstrated for \textit{image} gradients).
\textbf{Right: Qualitative examples on VOS.} \oursName reveals finer details of the semantic masks compared to existing upsampling strategies, such as FeatUp~\cite{Fu:2024:FeatUp}.
}
\label{fig:edge_loss}
\vspace{-0.8em}
\end{figure}
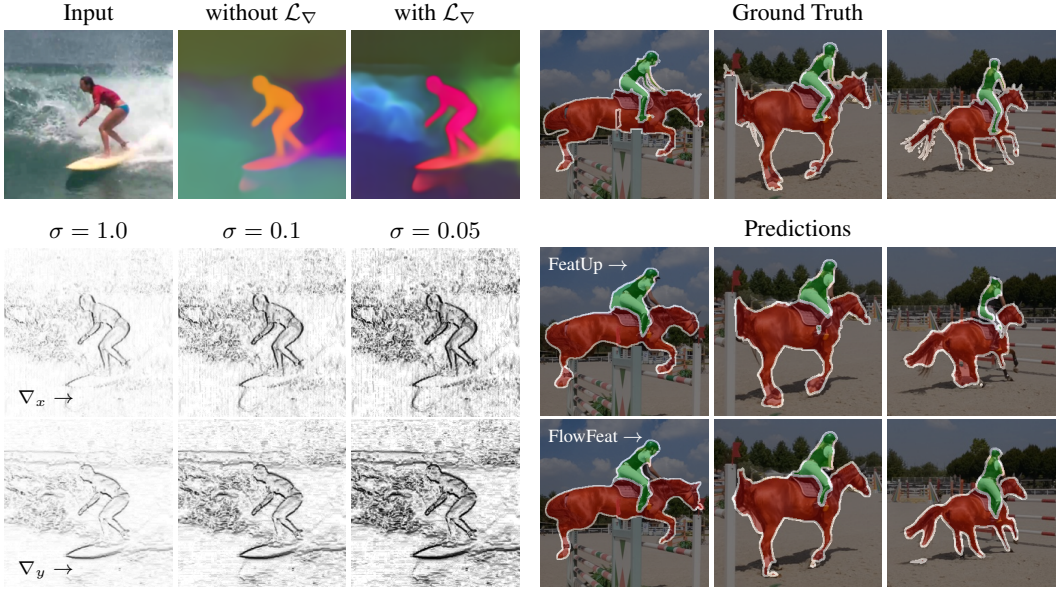

\cref{fig:edge_loss} illustrates the effect of the gradient matching loss.
As we also demonstrate empirically in \cref{sec:ablation}, the gradient loss results in sharper feature maps (see the top row in \cref{fig:edge_loss}).
Note that the focal term in \cref{eq:edge} enables modulation of the gradient loss at motion discontinuities.
As the two bottom rows in \cref{fig:edge_loss} demonstrate, the hyperparameter $\sigma$ controls the degree of this modulation: a lower value of $\sigma$ results in sharper \oursName boundaries.
However, a very low value of $\sigma$ may amplify the negative effect of inaccurate flow predictions, which can also exhibit flow discontinuities.

The total second-order flow reconstruction loss is simply a weighted sum:
\begin{equation}\label{eq:main_loss}
    \mathcal{L}_\text{total} = \mathcal{L}_\nabla + \lambda \mathcal{L}_{L1},
\end{equation}
where $\mathcal{L}_\nabla$ is the sum of $\mathcal{L}^x_\nabla$ and $\mathcal{L}^y_\nabla$, and $\lambda$ is a trade-off hyperparameter kept fixed across all models.

%% file: figures/framework.tex
\begingroup%
  \makeatletter%
  \providecommand\color[2][]{%
    \errmessage{(Inkscape) Color is used for the text in Inkscape, but the package 'color.sty' is not loaded}%
    \renewcommand\color[2][]{}%
  }%
  \providecommand\transparent[1]{%
    \errmessage{(Inkscape) Transparency is used (non-zero) for the text in Inkscape, but the package 'transparent.sty' is not loaded}%
    \renewcommand\transparent[1]{}%
  }%
  \providecommand\rotatebox[2]{#2}%
  \newcommand*\fsize{\dimexpr\f@size pt\relax}%
  \newcommand*\lineheight[1]{\fontsize{\fsize}{#1\fsize}\selectfont}%
  \ifx\svgwidth\undefined%
    \setlength{\unitlength}{127.73991797bp}%
    \ifx\svgscale\undefined%
      \relax%
    \else%
      \setlength{\unitlength}{\unitlength * \real{\svgscale}}%
    \fi%
  \else%
    \setlength{\unitlength}{\svgwidth}%
  \fi%
  \global\let\svgwidth\undefined%
  \global\let\svgscale\undefined%
  \makeatother%
  \begin{picture}(1,0.61)%
    \footnotesize
    \lineheight{1}%
    \setlength\tabcolsep{0pt}%
    \put(0,0){\includegraphics[width=\unitlength,page=1]{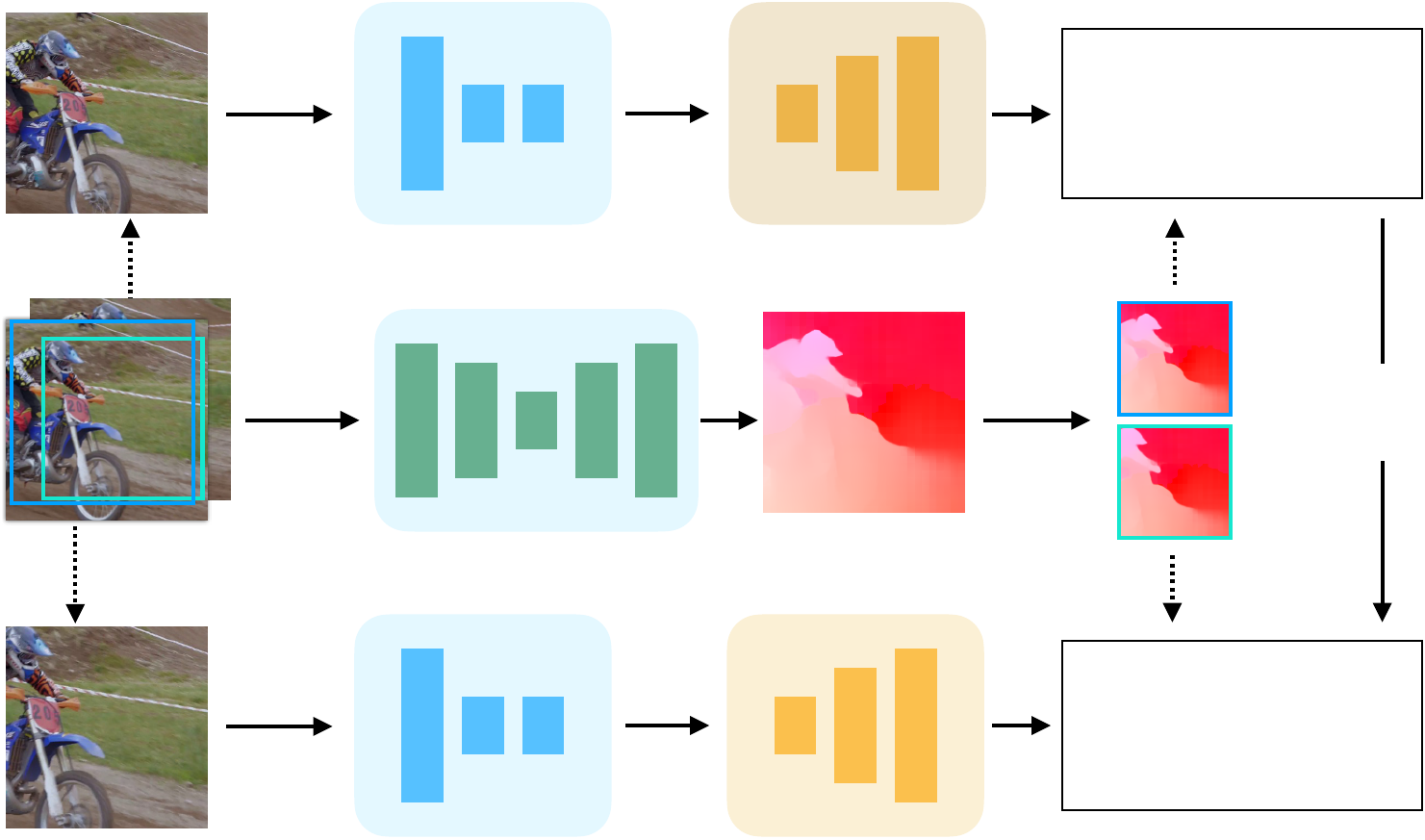}}%
    \put(0.181,0.545){\color[rgb]{0,0,0}\makebox(0,0)[lt]{$v_1$}}%
    \put(0.181,0.12){\color[rgb]{0,0,0}\makebox(0,0)[lt]{$v_2$}}%
    \put(0.168,0.339){\color[rgb]{0,0,0}\makebox(0,0)[lt]{$I_t, I_{t^\prime}$}}%
    \put(0.335,0.575){\color[rgb]{0,0,0}\makebox(0,0)[lt]{$\mathcal{E}$}}%
    \put(0.335,0.145){\color[rgb]{0,0,0}\makebox(0,0)[lt]{$\mathcal{E}$}}%
    \put(0.36,0.36){\color[rgb]{0,0,0}\makebox(0,0)[lt]{$\mathcal{F}$}}%
    \put(0.53,0.575){\color[rgb]{0,0,0}\makebox(0,0)[lt]{$\mathcal{D}^\text{EMA}$}}%
    \put(0.53,0.145){\color[rgb]{0,0,0}\makebox(0,0)[lt]{$\mathcal{D}_\theta$}}%
    \put(0.95,0.32){\color[rgb]{0,0,0}\makebox(0,0)[lt]{\large $A^\ast$}}%
    \put(0.84,0.55){\color[rgb]{0,0,0}\makebox(0,0)[lt]{$\text{min}_A$}}%
    \put(0.775,0.51){\color[rgb]{0,0,0}\makebox(0,0)[lt]{$\| u_1 - x_1 A\|$}}
    \put(0.84,0.12){\color[rgb]{0,0,0}\makebox(0,0)[lt]{$\text{min}_\theta$}}%
    \put(0.77,0.08){\color[rgb]{0,0,0}\makebox(0,0)[lt]{$\| u_2 - x_2 A^\ast\|$}}

    \put(0.875,0.27){\color[rgb]{0,0,0}\makebox(0,0)[lt]{$u_2$}}
    \put(0.875,0.35){\color[rgb]{0,0,0}\makebox(0,0)[lt]{$u_1$}}
    \put(0.699,0.54){\color[rgb]{0,0,0}\makebox(0,0)[lt]{$x_1$}}
    \put(0.699,0.115){\color[rgb]{0,0,0}\makebox(0,0)[lt]{$x_2$}}
    \put(0.59,0.408){\color[rgb]{0,0,0}\makebox(0,0)[lt]{$F$}}
  \end{picture}%
\endgroup%

%% file: sections/04_experiments.tex
\section{Experiments}
\label{sec:experiments}

We probe \oursName on three diverse tasks: video object segmentation (VOS), semantic segmentation and monocular depth prediction.
Our goal is to demonstrate that \oursName offers substantial and consistent benefits across these downstream tasks as well as across backbone models, regardless of their pre-training strategy.
Overall, we train \oursName on top of five backbone models: Masked Autoencoder (MAE) \cite{He:2022:MAA} based on ViT-B16, DINO \cite{Caron:2021:EPS} based on ViT-B16 and ViT-S16, and DINO2 \cite{Oquab:2023:DINOv2} based on ViT-B14 and ViT-S14.
As the decoder architecture and the only trainable component in \oursName, we use the DPT model \cite{Ranftl:2021:VTD}, which is runtime-efficient (\cf \cref{tab:complexity}, \supp).
The flow distillation relies on SEA-RAFT \cite{Wang:2024:SEA} based on ResNet-34.
However, our ablation experiments in \cref{sec:ablation} with the older RAFT model \cite{TeedD20} and unsupervised flow \cite{Stone:2021:SMURF} show that this choice of the flow estimator is not critical.
Furthermore, \cref{fig:flow_approx_err} illustrates the resilience of the training to inaccurate flow targets.
We report the results for two \oursName variants.
\oursName\!\!\textit{-YT} trains on $3471$ video sequences from YouTube-VOS (CC BY 4.0, \cite{Xu:2018:YTV}).
For larger backbones, we train \oursName\!\!\textit{-K} on Kinetics-400 dataset (CC BY 4.0, \cite{Kay:2017:KHA}) containing $147646$ videos.\footnote{We exclude videos containing a montage of multiple clips to ensure temporal coherence.}
We compare our \oursName variants to the corresponding encoder model, as well as FeatUp \cite{Fu:2024:FeatUp}, pre-trained on COCO-Stuff (CC BY 4.0 / Flickr, \cite{caesar2018cvpr}). Recall that FeatUp stacks multiple bilateral upsamplers and preserves the feature dimensionality.
For instance, FeatUp yields representations with dimensionality $384$ for ViT-S, whereas \oursName is more compact and has a fixed dimensionality of $128$ across all variants.
This allows us to evaluate \oursName in a complementary fashion to the backbone encoding by jointly fitting a high-resolution probe on \oursName and a low-resolution probe on the fixed encoder.

\inparagraph{Implementation details (see also \cref{supp-sec:impl_details}).}
Training \oursName is computationally inexpensive.
To train one model, we use a \emph{single} GPU with $46$GB of memory.
The training proceeds with mini-batches of $128$ images, input resolution $224 \times 224$ and AdamW optimiser \cite{Kingma:2015:AAM,Loshchilov:2019:DWD} with learning rate $10^{-4}$ and no weight decay.
For the hyperparameters, we empirically set $\lambda = 0.1$, $\sigma=0.1$ and $\gamma = 1.0$ and did not observe sensitivity to moderate deviations from these values.
We train \oursName for 500 epochs on YouTube-VOS and for 100 epochs on Kinetics.
In wall-clock time with one A40 GPU, the training takes only 24 hours and 3 days for YouTube-VOS and Kinetics, respectively.

\begin{table}[t]
\begin{minipage}[t]{.68\textwidth}
\newcolumntype{Y}{S[table-format=2.1]@{\hspace{1em}}} 
\newcolumntype{Z}{S[table-format=2.1]} 
\medskip
\scriptsize
\centering
\begin{tabularx}{\linewidth}{lXYYZl@{\hspace{1em}}YYZ}
\toprule
\multirow{2}{*}{Method} & \multirow{2}{*}{Train Data} & \multicolumn{3}{c}{Linear Probing} & & \multicolumn{3}{c}{Local KNN} \\
\cmidrule(lr){3-5} \cmidrule(lr){7-9}
& & $\mathcal{JF}$ & $\mathcal{J}_m$ & $\mathcal{F}_m$ & & $\mathcal{JF}$ & $\mathcal{J}_m$ & $\mathcal{F}_m$ \\
\midrule
V-JEPA~\cite{Bardes:2024:RFP} & VideoMix2M~\cite{Bardes:2024:RFP}   & 49.0 & 46.1 & 51.9  & & 56.7 & 55.6 & 57.8 \\
\midrule
VideoMAE~\cite{tong:2022:videomae} & Kinetics   & 43.3 & 40.9 & 45.8 & & 55.1 & 54.6 & 55.6 \\
\midrule
MAE-B16~\cite{He:2022:MAA} & ImageNet    & 40.8 & 38.5 & 43.1 & & 44.3 & 42.8 & 45.8 \\
\rowcolor{azure}
\leftPad \oursName-K           & \leftPad  Kinetics & \bfseries 53.8 & \bfseries 50.1 & \bfseries 57.5 & & \bfseries 59.1 & \bfseries 57.3 & \bfseries 60.8 \\ 
\midrule
DINO-B16~\cite{Caron:2021:EPS} & ImageNet         & 52.3 & 49.1 & 55.4 & & 62.3 & 60.7 & 64.0 \\
\rowcolor{azure}
\leftPad \oursName-YT           & \leftPad  YT-VOS & 55.5 & 52.5 & 58.5 & & 64.0 & 62.7 & 65.3 \\ 
\rowcolor{azure}
\leftPad \oursName-K           & \leftPad  Kinetics & \bfseries 56.9 & \bfseries 53.7 & \bfseries 60.1 & & \bfseries 66.0 & \bfseries 64.5 & \bfseries 67.5 \\ 
\midrule
DINO-S16 \cite{Caron:2021:EPS}    &  ImageNet        & 49.6 & 46.8 & 52.4 & & 61.5 & 59.9 & 63.1 \\
\leftPad FeatUp~\cite{Fu:2024:FeatUp} &   COCO-S  & 52.4 & 49.6 & 55.2 & & 63.7 & 62.4 & 64.9 \\
\rowcolor{azure}
\leftPad \oursName-YT              & \leftPad  YT-VOS & 54.1 & 51.1 & 57.0 & & 63.7 & 62.0 & 65.5 \\ 
\rowcolor{azure}
\leftPad \oursName-K              & \leftPad  Kinetics & \bfseries 56.2 & \bfseries 52.9 & \bfseries 59.5 & & \bfseries 66.5 & \bfseries 64.5 & \bfseries 68.4 \\ 
\midrule
DINO2-B14 \cite{Oquab:2023:DINOv2} &              LVD$^\ast$        & 61.6 & 58.5 & 64.7 & & 66.4 & 64.4 & 68.5 \\
\rowcolor{azure}
\leftPad \oursName-YT                & \leftPad  YT-VOS     & 65.7 & 62.2 & 69.2 & & 69.0 &  66.9 & 71.2 \\ 
\rowcolor{azure}
\leftPad \oursName-K                &  \leftPad  Kinetics  & \bfseries 66.1 & \bfseries 62.3 & \bfseries 69.9 & & \bfseries 69.9 & \bfseries 67.7 & \bfseries 72.1 \\
\midrule
DINO2-S14 \cite{Oquab:2023:DINOv2} &    LVD$^\ast$ & 57.5 & 54.2 & 60.7 & & 65.1 & 63.7 & 66.6 \\
\leftPad FeatUp~\cite{Fu:2024:FeatUp}   & COCO-S    & 60.5 & 57.4 & 63.6 & & 65.5 & 65.0 & 66.1 \\
\leftPad LoftUp~\cite{Haiwen:2025:LUP}        & \leftPad   SA1B \cite{kirillov:2023:SAM}     & 63.0 & 59.6 & 66.4 &  & 66.0  & 64.7  & 67.4 \\ 
\rowcolor{azure}
\leftPad \oursName-YT               & \leftPad  YT-VOS    & \bfseries 65.8 & \bfseries 62.0 & \bfseries 69.7 & & 67.6 & 65.6 & 69.6\\ 
\rowcolor{azure}
\leftPad \oursName-K                & \leftPad  Kinetics    & 64.6 & 61.0 & 68.2 &  & \bfseries 68.5 & \bfseries 66.1 & \bfseries70.9 \\ 
\bottomrule
\end{tabularx}
\end{minipage}
\hfill
\begin{minipage}[t]{.3\textwidth}
    \vspace{-0.3em}
    \caption{\textbf{Video object segmentation (VOS) with linear probing and label propagation (local KNN) on DAVIS-2017 (val).} \oursName significantly improves the VOS accuracy of the baselines across all tested scenarios. It further outperforms previous and concurrent upsampling techniques (FeatUp~\cite{Fu:2024:FeatUp} and LoftUp~\cite{Haiwen:2025:LUP}). Pre-training \oursName on the larger Kinetics datasets tends to produce a stronger representation. LVD$^\ast$ refers to the distillation from a model pre-trained on LVD \cite{Oquab:2023:DINOv2}. LoftUp~\cite{Haiwen:2025:LUP} uses SAM, trained with mask supervision on SA1B \cite{kirillov:2023:SAM}.}
\label{tab:vos_results}
\end{minipage}
\vspace{-0.8em}
\end{table}

\subsection{Video object segmentation}
\label{sec:eval_vos}

We evaluate \oursName on semi-supervised video object segmentation (VOS) using $30$ validation sequences from DAVIS-2017 (CC BY-SA 4.0, \cite{PontTuset2017}).
The task is to propagate the ground-truth annotation defined in the first frame to the rest of the video.
Therefore, performing well on this task would indicate the capacity for temporal invariance as well as pixel-level semantic discrimination.

Previous evaluation protocols for VOS employ a variant of a localised k-nearest neighbour classifier \cite{Jabri:2020:STC,Lai:2020:MAS,Araslanov:2021:DUL}, referred to as \emph{local KNN} in the following.
This probing technique is known to be brittle, exhibiting high volatility \wrt its hyperparameters \cite{McKee:2022:TRV}.
For consistency with previous work, we stick to the implementation of local KNN provided by \citet{Caron:2021:EPS}.
However, we additionally evaluate VOS with \emph{linear probing}, as the more established and interpretable technique in representation learning \cite{Chen:2020:ASF,GrillSATRBDPGAP20}.
Linear probing extends seamlessly to the VOS task.
Specifically, for each video, we train a linear classifier using the ground-truth segmentation provided for the first frame.
We apply the linear classifier to the remaining frames to obtain the segmentation result.
For both probing strategies -- linear probing and local KNN -- we compute the mean region similarity $\mathcal{J}_m$, the mean contour-based accuracy $\mathcal{F}_m$ and their mean $\mathcal{JF}$.
\cref{tab:vos_results} reports the results. 
Across all pre-training methods and metrics, \oursName achieves a consistent and substantial improvement in VOS accuracy.
The benefit is especially significant for MAE-B16, where \oursName improves the baseline by staggering $13.0\%$ / $14.8\%$ \JF with linear probing / local KNN.
However, \oursName also surpasses stronger baselines, \eg DINO2-B14 ($+4.5 \%$ / $+3.5 \%$ \JF) and FeatUp ($+3.8 \%$ / $+2.8 \%$ \JF for DINO-S16 and $+5.3 \%$ / $+3.0 \%$ for DINO2-S14 \JF). 
As illustrated in \cref{fig:edge_loss} (right), the improvement is especially pronounced at the object boundaries.
FeatUp enhances VOS accuracy for both baselines (DINO-S16 and DINO2-S14), but these improvements are more modest.
FeatUp also struggles with inputs of higher resolution, introducing static feature artefacts, see the supplemental videos and further analysis in \cref{supp-tab:ablation_knn}.
Similarly, the contemporaneous work, LoftUp~\cite{Haiwen:2025:LUP}, achieved inferior accuracy despite the implicit leverage of vast mask supervision via SAM \cite{kirillov:2023:SAM}.
Video-based models, such as V-JEPA~\cite{Bardes:2024:RFP} and VideoMAE~\cite{tong:2022:videomae}, are also remarkably ineffective.

Overall, the improvements on VOS metrics provide compelling evidence that \oursName encapsulates a high degree of temporal invariance and feature detail, with complementary properties to the encoder representation.
Furthermore, the larger Kinetics dataset tends to produce a stronger variant of \oursName.
This observation indicates that \oursName has the promising capacity to scale with the ever-increasing volume of real-world videos.

\begin{table}[t]

\captionof{table}{\textbf{Probing semantic segmentation and monocular depth.} On COCO-Stuff 2017 (val), \oursName advances the segmentation quality across all baselines as well. A lightweight refinement using \oursName{++} (numbers in parentheses) further boosts the accuracy without any parameter training. On NYUv2 (val), \oursName significantly improves the depth accuracy across all pre-trained encoders -- in contrast to FeatUp, which struggles to improve upon its baselines. }
\label{tab:depth_results}

\newcolumntype{B}{S[table-format=1.4]}
\newcolumntype{A}{S[table-format=2.2]}
\newcolumntype{Z}{S[table-format=2.2]} 
\newcolumntype{Y}{S[table-format=2.1]} 
\newcolumntype{Q}{S[table-format=2.1,input-symbols = ()]} 

\medskip
\scriptsize
\centering
\begin{tabularx}{\linewidth}{lXYQYQlBAAZ}
\toprule
\multirow{2}{*}{\textbf{Method}} & & \multicolumn{4}{c}{Semantic Segmentation} & &  \multicolumn{4}{c}{Depth Estimation} \\
\cmidrule(lr){3-6} \cmidrule(lr){8-11}
 & & \multicolumn{2}{c}{mIoU $\uparrow$} & \multicolumn{2}{c}{pAcc $\uparrow$} & & {RMSE $\downarrow$} & {$\delta_1 \uparrow$} & {$\delta_2 \uparrow$} & {$\delta_3 \uparrow$}  \\
\midrule
MAE-B16 \cite{He:2022:MAA} &       & 46.0 & & 71.5 & &     & 0.4534 & 83.68 & 96.98 & 99.28 \\
\rowcolor{azure}
\leftPad\oursName-K           & & \bfseries 47.2 & & \bfseries 72.9 & &  & \bfseries 0.4400 & \bfseries 84.43 & \bfseries 97.18 & \bfseries 99.35   \\ 
\midrule
DINO-B16 \cite{Caron:2021:EPS} & & 46.1 & & 72.0 &  &           & 0.4287 & 86.15 & 97.61 & 99.47 \\
\rowcolor{azure}
\leftPad\oursName-K        & & \bfseries 48.2 & & \bfseries 73.7 &     &  & \bfseries 0.4176 & \bfseries 86.87 & \bfseries 97.71 & \bfseries 99.50 \\ 
\midrule
FeatUp -- DINO-S16~\cite{Fu:2024:FeatUp}~(++) & & 41.6 & (42.1) & 69.5 & (69.9)  &          & 0.4624 & 83.54 & 96.90 & 99.32\\
DINO-S16 \cite{Caron:2021:EPS}   & & 39.6 & & 67.5 &  &  & 0.4634 & 83.60 & 96.94 & 99.32  \\
\rowcolor{azure}
\leftPad\oursName-YT~(++)           & & \bfseries 44.7 & \bfseries (45.9) & \bfseries 71.4 & \bfseries  (72.5)    & & \bfseries 0.4410 & \bfseries 85.26 & 97.17 & 99.30  \\ 
\rowcolor{azure}
\leftPad\oursName-K~(++)        & & 44.2 & (45.4) & 71.3 & (72.3)       & &  0.4422 & 84.81 & \bfseries 97.19 & \bfseries 99.37 \\ 
\midrule
DINO2-B14 \cite{Oquab:2023:DINOv2}  & & 58.1 & & 78.0 & & & 0.3091 & 94.14 & 99.32 & 99.89 \\
\rowcolor{azure}
\leftPad\oursName-K          & & \bfseries 60.4 & & \bfseries 79.8 &        &  & \bfseries  0.2791 & \bfseries  95.55 & \bfseries 99.52 & \bfseries 99.93 \\
\midrule
FeatUp -- DINO2-S14\cite{Fu:2024:FeatUp}~(++)   & & \bfseries 58.3 & (58.5) & \bfseries 79.1 & (79.2)  &   & 0.3207 & 93.29 & 99.18 & 99.86\\ 
DINO2-S14 \cite{Oquab:2023:DINOv2}  & & 56.2 & & 77.3 &  &   & 0.3294 & 92.97 & 99.11 & 99.85 \\
\rowcolor{azure}
\leftPad\oursName-YT~(++)         & & 58.0 &  (59.4) & 78.7 & (79.7)       & & 0.3072 & 93.91 & 99.25 & 99.86  \\ 
\rowcolor{azure}
\leftPad\oursName-K~(++)       & & 58.1 & \bfseries  (59.6) & 78.9 & \bfseries  (79.9)          & & \bfseries 0.3061 & \bfseries 94.12 & \bfseries 99.31 & \bfseries 99.88 \\ 
\bottomrule
\end{tabularx}
\vspace{-0.8em}
\end{table}

\subsection{Semantic segmentation}

We follow the setting of FeatUp \cite{Fu:2024:FeatUp} and use COCO-Stuff 2017 with $C=27$ coarsely annotated categories \cite{caesar2018cvpr}.
Since \oursName focuses on motion patterns rather than global semantic alignment, it may lack consistent semantic structure across images; therefore, we employ attention probing \cite{Bardes:2024:RFP} to derive image-specific class prototypes.
In more detail, we define $C=27$ learnable queries attending the \oursName representation with a single layer of cross-attention. 
We freeze the models and train the probes on $256 \times 256$ centre crops using the cross-entropy loss. 
Additionally, we demonstrate that \oursName can further boost the segmentation accuracy with a simple adaptation of a lightweight post-processing technique.
Concretely, we adapt the local mask refinement strategy (PAMR) \cite{Araslanov:2020:SSS}, but leverage \oursName instead of the image intensity to refine the segmentation result.
Note that such a refinement is not possible by the use of the probes alone due to their feed-forward nature.
We refer to this straightforward extension as \oursNamePP.
\cref{supp-sec:impl_details_seg} provides further details.

\cref{tab:depth_results} reports the mean pixel accuracy and the mean IoU.
The results align with our observations in VOS experiments: \oursName boosts the accuracy across all baseline models.
Particularly notable are the improvements \wrt smaller models.
For example, \oursName surpasses DINO-S16 by $5.1 \%$ and $4.6 \%$ with \oursNameX{YT} and \oursNameX{K}, respectively.
Without the refinement, \oursName performs competitively with FeatUp \cite{Fu:2024:FeatUp} based on DINO2-S14 and outperforms it for DINO-S16.
Furthermore, the \oursName-based refinement significantly enhances the segmentation quality.
For example, \oursNameXPP{K} improves over \oursNameX{K} by a notable margin of $1.5 \%$ mIoU.
By contrast, FeatUp does not profit from the refinement as much.

\begin{figure*}[t]
\captionsetup[subfigure]{labelformat=empty}
\centering

\begin{subfigure}{0.161\linewidth}
\centering
\scriptsize{Input}
\end{subfigure}
\begin{subfigure}{0.161\linewidth}
\centering
\scriptsize{DINO2-S14}
\end{subfigure}
\begin{subfigure}{0.161\linewidth}
\centering
\scriptsize{\oursName-K}
\end{subfigure}
\begin{subfigure}{0.161\linewidth}
\centering
\scriptsize{\oursName-K (PCA)}
\end{subfigure}
\begin{subfigure}{0.161\linewidth}
\centering
\scriptsize{\oursName-K++}
\end{subfigure}
\begin{subfigure}{0.161\linewidth}
\centering
\scriptsize{Ground Truth}
\end{subfigure}
\vspace{0.1em}
\begin{subfigure}{0.161\linewidth}
\includegraphics[width=\linewidth]{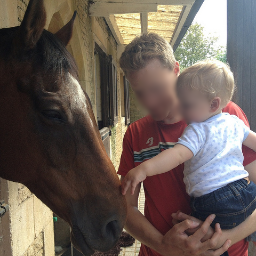}
\end{subfigure}
\begin{subfigure}{0.161\linewidth}
\includegraphics[width=\linewidth]{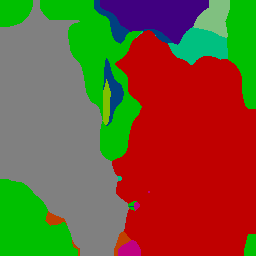}
\end{subfigure}
\begin{subfigure}{0.161\linewidth}
\includegraphics[width=\linewidth]{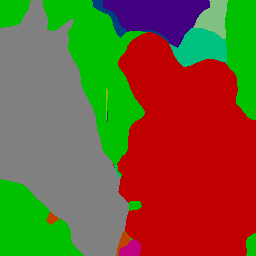}
\end{subfigure}
\begin{subfigure}{0.161\linewidth}
\includegraphics[width=\linewidth]{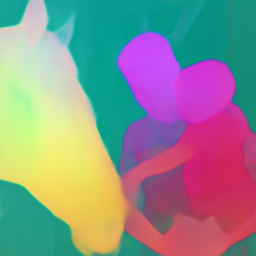}
\end{subfigure}
\begin{subfigure}{0.161\linewidth}
\includegraphics[width=\linewidth]{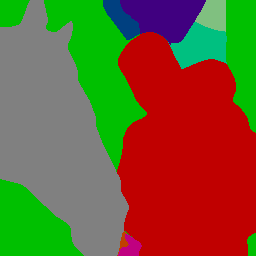}
\end{subfigure}
\begin{subfigure}{0.161\linewidth}
\includegraphics[width=\linewidth]{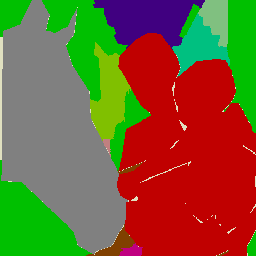}
\end{subfigure}

\begin{subfigure}{0.161\linewidth}
\includegraphics[width=\linewidth]{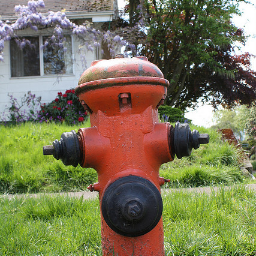}
\end{subfigure}
\begin{subfigure}{0.161\linewidth}
\includegraphics[width=\linewidth]{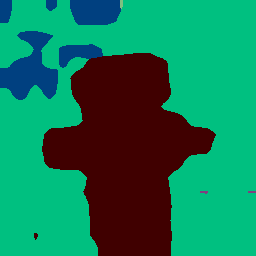}
\end{subfigure}
\begin{subfigure}{0.161\linewidth}
\includegraphics[width=\linewidth]{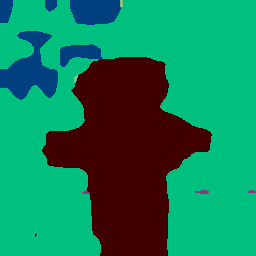}
\end{subfigure}
\begin{subfigure}{0.161\linewidth}
\includegraphics[width=\linewidth]{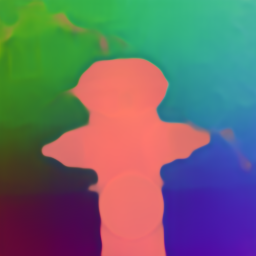}
\end{subfigure}
\begin{subfigure}{0.161\linewidth}
\includegraphics[width=\linewidth]{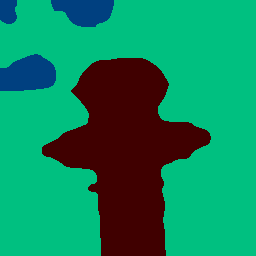}
\end{subfigure}
\begin{subfigure}{0.161\linewidth}
\includegraphics[width=\linewidth]{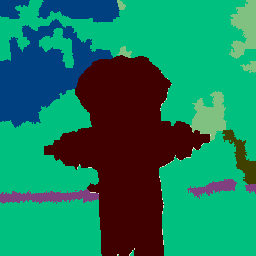}
\end{subfigure}

\caption{\textbf{Semantic segmentation and post-hoc refinement (++) with \oursName.} The segmentation masks from \oursName exhibit a high level of boundary accuracy. The \oursName representation, visualised with PCA, identifies prominent scene elements with a fine-grained detail. A lightweight post-hoc refinement (\oursName-K++), based on PAMR \cite{Araslanov:2020:SSS}, leverages the pairwise pixel similarity embedded by \oursName (instead of image intensities) to improve the results further.}
\label{fig:seg_results}
\vspace{-0.5em}
\end{figure*}

\cref{fig:seg_results} visualises the segmentation results for the DINO2-S14 backbone, with and without the refinement.
Initial predictions of the probes are coarse and lack detail, especially around object boundaries.
Leveraging the high-resolution \oursName representation (visualised with PCA), the refinement leads to sharper mask alignment with image boundaries.

In summary, \oursName provides a significant boost also for downstream semantic tasks.
The feature representation offers a high degree of spatial detail and also lends itself well to lightweight post-processing without the need for additional training.

\subsection{Monocular depth estimation}
\label{sec:monodepth}

\begin{wrapfigure}[31]{r}{0.4\textwidth}
\vspace{-1.1em}
\captionsetup[subfigure]{labelformat=empty}
\centering
\scriptsize

\begin{subfigure}{0.326\linewidth}
\begin{picture}(0,0)
        \put(0,0){\includegraphics[width=\linewidth]{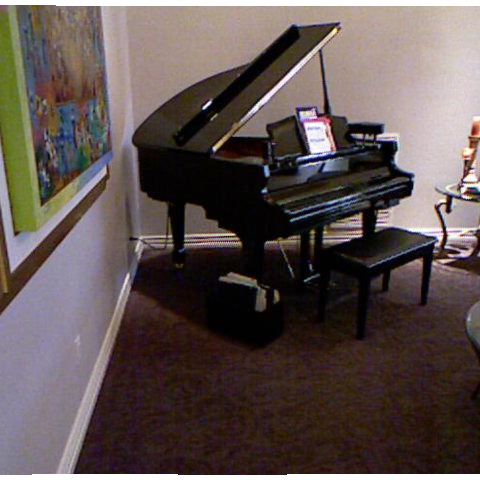}}
        \put(3,5){\scriptsize\color{white}{Input $\rightarrow$}}
    \end{picture}
\end{subfigure}
\begin{subfigure}{0.326\linewidth}
\includegraphics[width=\linewidth]{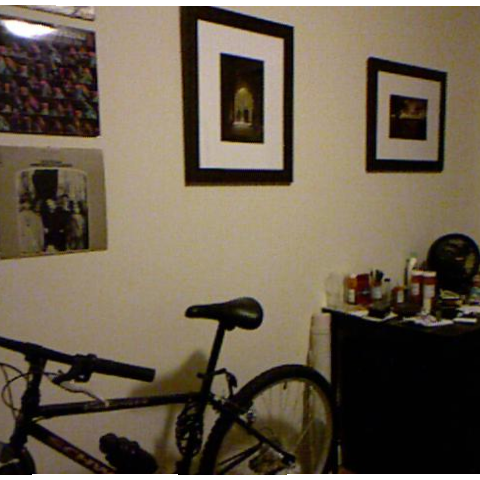}
\end{subfigure}
\begin{subfigure}{0.326\linewidth}
\includegraphics[width=\linewidth]{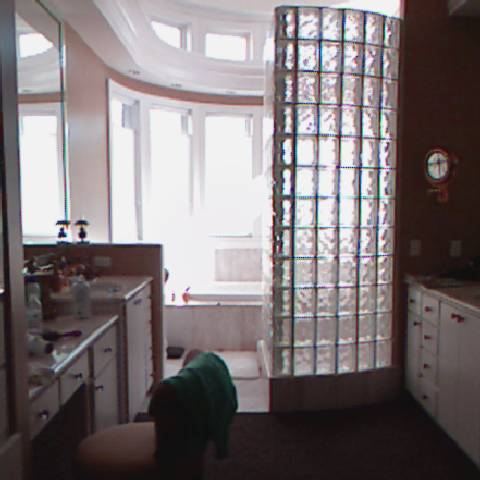}
\end{subfigure}

\begin{subfigure}{0.326\linewidth}
\begin{picture}(0,0)
        \put(0,0){\includegraphics[width=\linewidth]{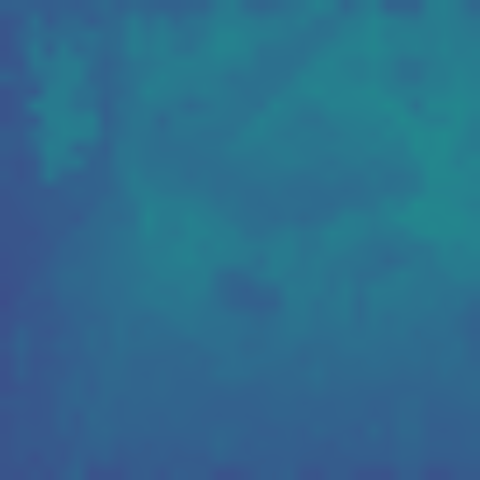}}
        \put(3,5){\scriptsize\color{white}{DINO2-S14}}
    \end{picture}
\end{subfigure}
\begin{subfigure}{0.326\linewidth}
\includegraphics[width=\linewidth]{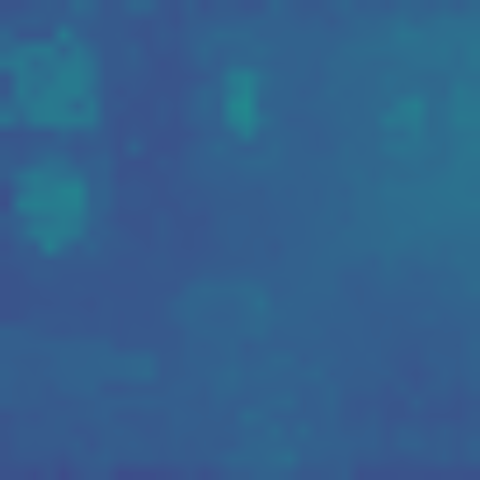}
\end{subfigure}
\begin{subfigure}{0.326\linewidth}
\includegraphics[width=\linewidth]{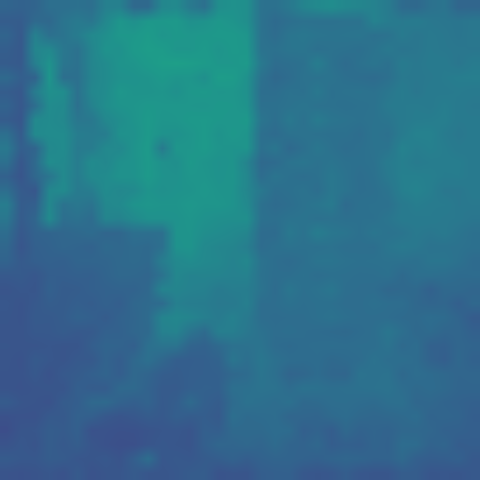}
\end{subfigure}

\begin{subfigure}{0.326\linewidth}
\begin{picture}(0,0)
        \put(0,0){\includegraphics[width=\linewidth]{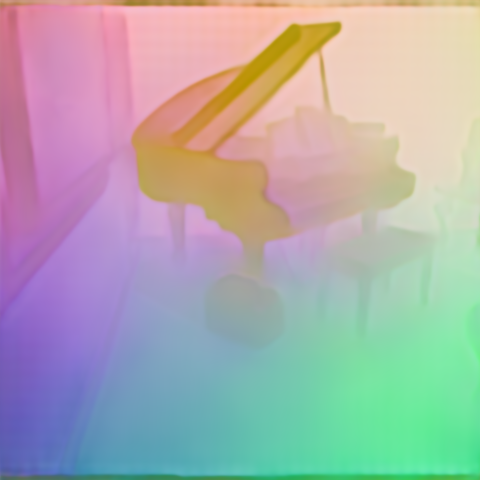}}
        \put(3,5){\scriptsize\color{white}{FlowFeat (PCA)}}
    \end{picture}
\end{subfigure}
\begin{subfigure}{0.326\linewidth}
\includegraphics[width=\linewidth]{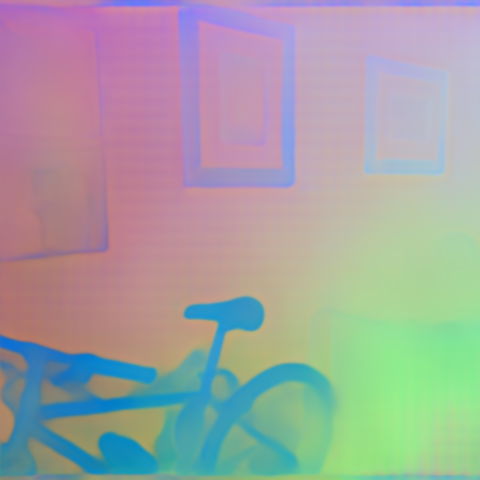}
\end{subfigure}
\begin{subfigure}{0.326\linewidth}
\includegraphics[width=\linewidth]{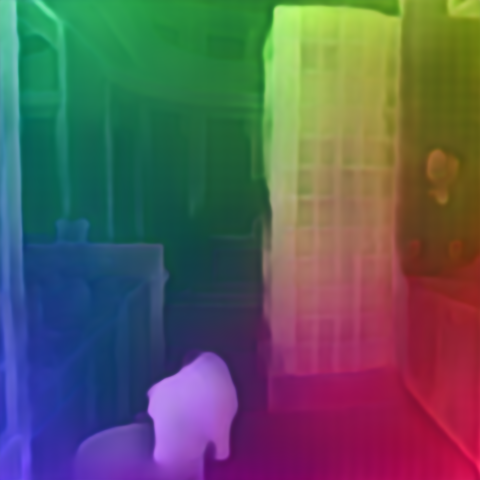}
\end{subfigure}

\begin{subfigure}{0.326\linewidth}
\begin{picture}(0,0)
        \put(0,0){\includegraphics[width=\linewidth]{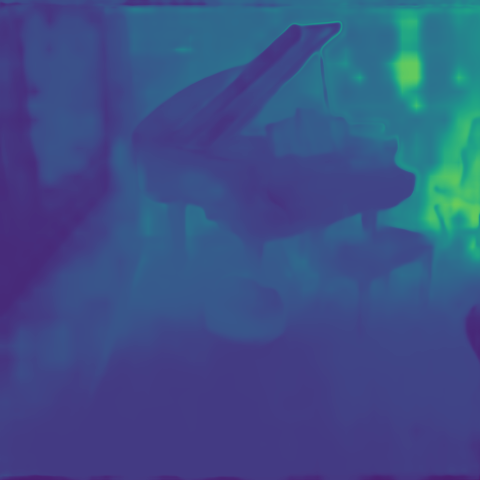}}
        \put(3,5){\scriptsize\color{white}{FlowFeat}}
    \end{picture}
\end{subfigure}
\begin{subfigure}{0.326\linewidth}
\includegraphics[width=\linewidth]{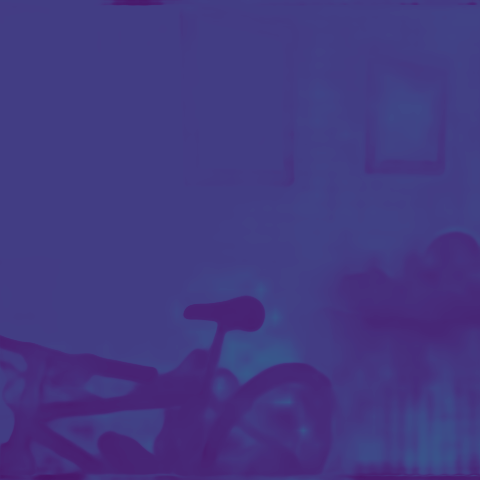}
\end{subfigure}
\begin{subfigure}{0.326\linewidth}
\includegraphics[width=\linewidth]{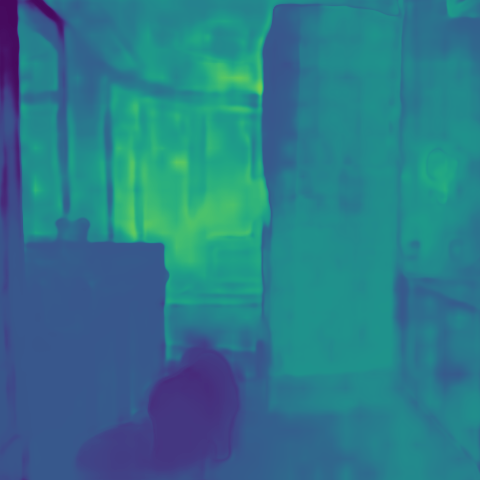}
\end{subfigure}

\begin{subfigure}{0.326\linewidth}
\begin{picture}(0,0)
        \put(0,0){\includegraphics[width=\linewidth]{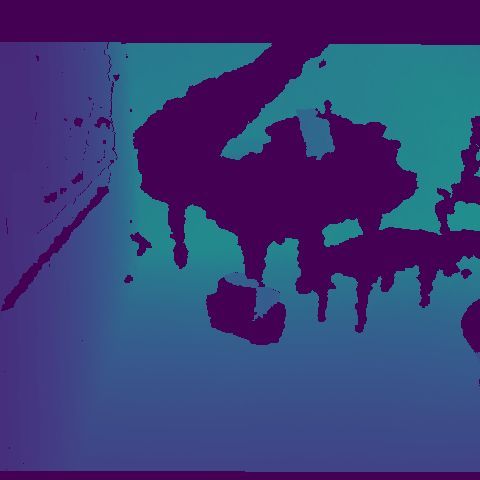}}
        \put(3,5){\scriptsize\color{white}{Ground Truth}}
    \end{picture}
\end{subfigure}
\begin{subfigure}{0.326\linewidth}
\includegraphics[width=\linewidth]{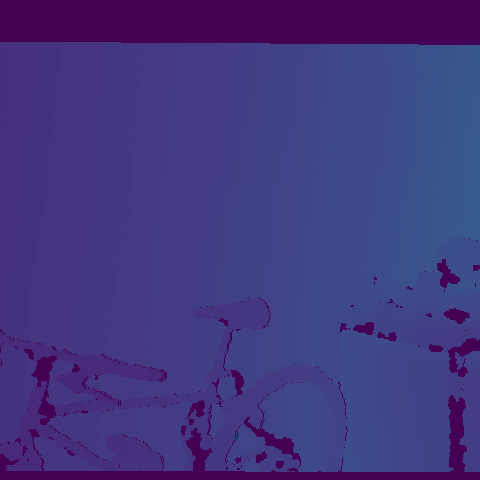}
\end{subfigure}
\begin{subfigure}{0.326\linewidth}
\includegraphics[width=\linewidth]{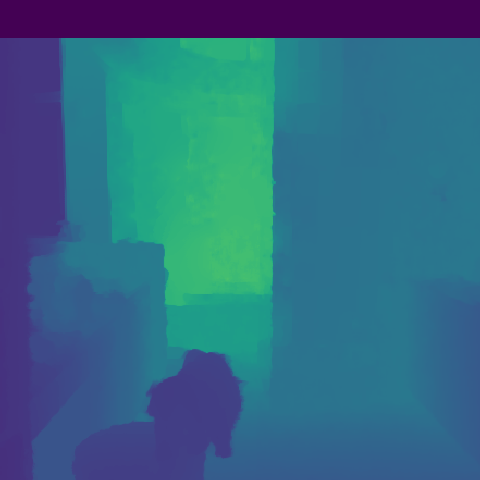}
\end{subfigure}

\caption{\textbf{Depth probing.} \oursName significantly improves depth estimates for challenging elements, such as non-Lambertian surfaces (\eg left, the piano), intricate structures (\eg middle, the bicycle), and under- and oversaturated image areas (\eg right, a bathroom).
}
\label{fig:depth_results}
\end{wrapfigure}
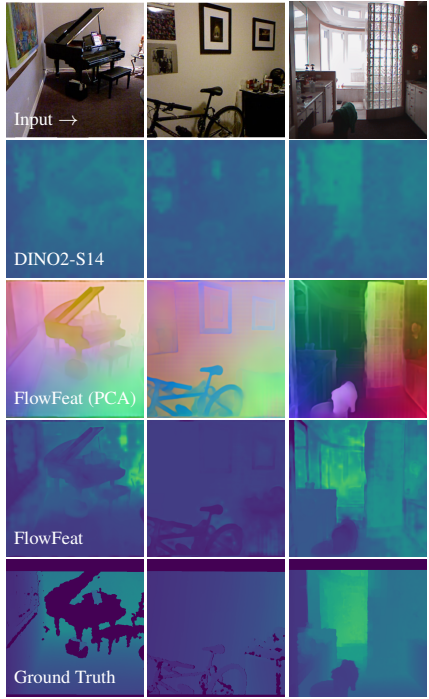
We evaluate \oursName on a geometric task, monocular depth estimation, using NYUv2 \cite{Silberman:ECCV12}, following the evaluation protocol of \citet{Banani:2024:P3D}.
Similar to the VOS setting, we compare \oursName against self-supervised backbones: MAE \cite{He:2022:MAA}, DINO \cite{Caron:2021:EPS}, and DINO2 \cite{Oquab:2023:DINOv2}.
As in semantic segmentation, we use attention probing \cite{Bardes:2024:RFP} to extract the depth-specific prototypes from \oursName. 
Specifically, we utilise the AdaBins \cite{Bhat2020AdaBinsDE} formulation that quantises the depth into 256 bins.
The depth value is a weighted sum of the predicted distribution across the bins and the corresponding depth value of the bin. Following \citet{Banani:2024:P3D}, we optimise the model using a weighted combination of the scale-invariant depth loss \cite{Bhat2020AdaBinsDE} and a gradient loss. 
\cref{supp-sec:impl_details} provides further details on probe implementation and training.

Adhering to the setting of \citet{Banani:2024:P3D}, we train the probes on the NYUv2's training set (24231 images) and evaluate the models on $480 \times 480$ centre crops
of the 1449 validation images \cite{Silberman:ECCV12}. As the standard depth metrics, we report the root-mean-squared error (RMSE) and the inlier rates at three thresholds.
Specifically, $\delta_i$ corresponds to the fraction of depth predictions $d$ satisfying $\text{max}(d / d^\ast, d^\ast / d) < 1.25^i$ \wrt the ground-truth $d^\ast$.

\cref{tab:depth_results} summarises the quantitative results.
In line with our observations on VOS and semantic segmentation, \oursName achieves a notable boost across all baseline models.
For example, \oursName-K reduces RMSE \wrt the DINO-S16 model by $0.051$ and increases the $\delta_1$ by $3.16 \%$.
By contrast, we did not observe benefit from the high-resolution FeatUp, which appears to be biased towards the pre-training resolution of $224 \times 224$.
Notably, we did not observe such a detrimental bias in \oursName.

We visually inspect the results in \cref{fig:depth_results} for the DINO2-S14 backbone.
In contrast to the low-quality depth estimates extracted from the frozen encoder, \oursName representation exhibits an impressive degree of fine-grained detail.
This is indeed noteworthy, considering that \oursName originates from video data and was not trained for such static scenes.
\oursName is also robust to under- and oversaturated image areas (\cf \cref{fig:depth_results}, the rightmost column) and infers highly plausible depth where the ground truth is not available due to surface specularity (\cf \cref{fig:depth_results}, the piano).

In summary, the results suggest that the motion profiles embedded by \oursName provide strong geometric cues.
\oursName not only enhances the depth awareness across all baselines, but also scales compellingly with the increased amount of video data for pre-training: \oursName-K outperforms the less data-intensive \oursName-YT across virtually all settings and metrics.

\begin{table}[t]
\begin{minipage}[t]{0.48\linewidth}
\caption{\textbf{Ablation study on DAVIS-2017 (val).} Following \cref{sec:eval_vos}, we perform linear probing on VOS in a set of ablation experiments. The $\Delta$ reports the absolute difference in the corresponding metric \wrt the baseline. The ridge regularisation in \oursName is crucial, but the choice of the flow estimator is not instrumental.}
\label{tab:ablation}
\newcolumntype{Y}{S[table-format=2.1]@{\hspace{0.3em}}} 
\newcolumntype{D}{S[table-format=+1.1]@{\hspace{1.3em}}} 
\newcolumntype{Z}{S[table-format=+1.1]} 
\medskip
\scriptsize
\centering
\begin{tabularx}{\linewidth}{@{}r@{\hspace{0.2em}}X@{\hspace{2em}}
S[table-format=2.1]@{\hspace{0.3em}}D
YD
YZ@{}}
\toprule
& {Baseline: DINO2-S14} & $\mathcal{JF}$ & $/\Delta$ &  $\mathcal{J}_m$ & {$/\Delta$} & $\mathcal{F}_m$ & $/\Delta$  \\
\midrule

& \textcolor{lightgrey}{+Random DPT}  &  \textcolor{lightgrey}{58.7} &  & \textcolor{lightgrey}{55.2} &  & \textcolor{lightgrey}{62.2} &  \\ 
& +\oursName-YT  &  \textbf{65.8} &  & \textbf{62.0} &  & \textbf{69.7} &  \\ 
\midrule 
\textit{(a)} & na\"{i}ve & 56.7 & \minus{9.1} & 52.8 & \minus{9.2} & 60.5 & \minus{9.2} \\ 
\textit{(b)} & $\gamma = 0.001$, \cf \cref{eq:ridge} & 58.2 & \minus{7.6} & 54.9 & \minus{7.1} & 61.5 & \minus{8.2} \\ 
\textit{(c)} & w/o $\mathcal{L}_\nabla$, \cf \cref{eq:edge} &  64.3 & \minus{1.5} & 61.0 & \minus{1.0} & 67.7 & \minus{2.0} \\ 
\textit{(d)} & $\mathcal{L}_\text{L2}$ &  63.3 & \minus{2.4} & 59.8 & \minus{2.2} & 67.0 & \minus{2.7} \\ 
\textit{(e)} & w/o $\mathcal{L}_\text{L1}$, \cf \cref{eq:loss_flow} &  65.3 & \minus{0.5} & 61.6 & \minus{0.4} & 69.0 & \minus{0.7} \\ 
\textit{(f)} & RAFT & 65.2 & \minus{0.6} & 61.6 & \minus{0.4} & 68.8 & \minus{0.9} \\ 
\textit{(g)} & SMURF (unsupervised) & 64.1 & \minus{1.7} & 60.7 & \minus{1.3} & 67.5 & \minus{2.2} \\
\textit{(h)} & temp. window $\times$ 2 & 65.5 & \minus{0.3} & 62.1 & \plus{0.1} & 68.9 & \minus{0.8} \\ 
\textit{(i)} & next frame only & 65.8 & 0.0 & 62.2 & \plus{0.2} & 69.4 & \minus{0.3} \\ 

\bottomrule
\end{tabularx}
\end{minipage}
\hfill
\begin{minipage}[t]{0.5\linewidth}
\vspace{1em}
\captionsetup[subfigure]{labelformat=empty}
\centering
\begin{subfigure}{0.24\linewidth}
\centering
\scriptsize{Input}
\end{subfigure}
\begin{subfigure}{0.24\linewidth}
\centering
\scriptsize{Target Flow}
\end{subfigure}
\begin{subfigure}{0.24\linewidth}
\centering
\scriptsize{Our Estimate}
\end{subfigure}
\begin{subfigure}{0.24\linewidth}
\centering
\scriptsize{\oursName (PCA)}
\end{subfigure}

\vspace{0.3em}

\begin{subfigure}{0.24\linewidth}
\includegraphics[width=\linewidth]{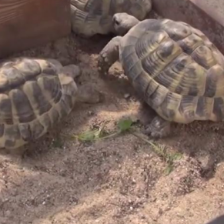}
\end{subfigure}
\begin{subfigure}{0.24\linewidth}
\includegraphics[width=\linewidth]{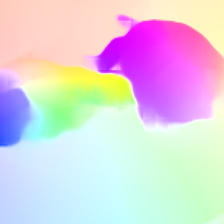}
\end{subfigure}
\begin{subfigure}{0.24\linewidth}
\includegraphics[width=\linewidth]{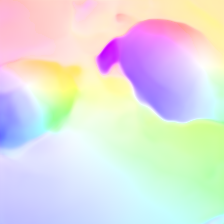}
\end{subfigure}
\begin{subfigure}{0.24\linewidth}
\includegraphics[width=\linewidth]{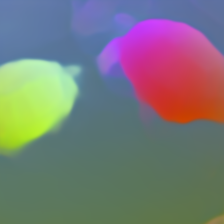}
\end{subfigure}

\vspace{0.5em}

\begin{subfigure}{0.24\linewidth}
\includegraphics[width=\linewidth]{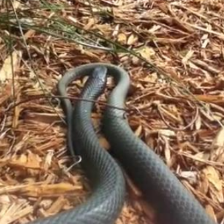}
\end{subfigure}
\begin{subfigure}{0.24\linewidth}
\includegraphics[width=\linewidth]{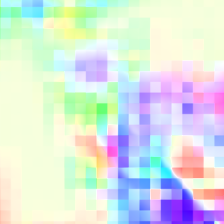}
\end{subfigure}
\begin{subfigure}{0.24\linewidth}
\includegraphics[width=\linewidth]{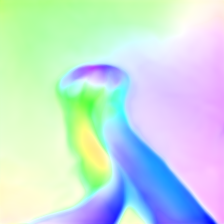}
\end{subfigure}
\begin{subfigure}{0.24\linewidth}
\includegraphics[width=\linewidth]{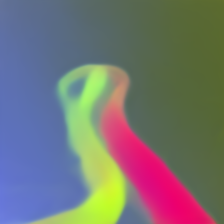}
\end{subfigure}

\vspace{0.8em}

\captionof{figure}{\textbf{Resilience to inaccurate flow targets.} Despite inaccurate and even artefact-prone predictions from optical flow networks, \oursName learns a reasonable flow approximation without compromising the feature representation.}
\label{fig:flow_approx_err}
\end{minipage}
\vspace{-0.5em}
\end{table}

\subsection{Ablation study}
\label{sec:ablation}

We conduct an ablation study of \oursName on the DAVIS-2017 (val) benchmark.
The study follows the evaluation protocol with linear probing from \cref{sec:eval_vos}.
Using ViT-S14 backbone pre-trained with DINO2 \cite{Oquab:2023:DINOv2}, we train a number of \oursName configurations on YouTube-VOS \cite{Xu:2018:YTV}.
\cref{tab:ablation} reports the results.
As a sanity check, we verify that a randomly initialised DPT decoder has virtually no effect on the VOS accuracy (\cf \cref{tab:ablation} in grey).
Similarly, \textit{(a)} na\"{i}vely fitting optical flow with a \emph{single} linear layer (instead of a distribution) is futile.
Next, \textit{(b)} we verify the benefit of estimating the optimal operator $A^\ast$ in \cref{eq:loss_flow} with ridge regression.
To compare with the baseline setting of $\gamma = 1$ (\cf \cref{eq:ridge}), we trained the model with $\gamma = 10^{-3}$.
We found the training numerically unstable with $\gamma = 0$.
Thus, setting $\gamma$ to $10^{-3}$ provides a reasonable approximation to removing the effect of L2-regularisation on the linear operator $A$.
In this case, the VOS accuracy drastically deteriorates across all metrics, which justifies the crucial need for ridge regularization.
\textit{(c)} We train \oursName without the second-order term, $\mathcal{L}_\nabla$ in \cref{eq:edge}.
A drop in the downstream accuracy (\eg $-2.0\%$ $\mathcal{F}_m$) suggests that \oursName exploits motion boundaries in its representation, in line with the established view that motion boundaries are strong semantic cues.
\textit{(d)} Replacing the $L_1$ reconstruction loss by $L_2$ distance in \cref{eq:loss_flow} reduces $\mathcal{J}\&\mathcal{F}$ from $65.8\%$ to $63.4\%$, supporting the robustness of $L_1$ in comparison to $L_2$.
Next, \textit{(e)} we explore a configuration without the L1 reconstruction term by setting $\lambda := 0$ in \cref{eq:main_loss}.
Surprisingly, the drop in accuracy is not substantial.
This suggests that the training process can succeed with the gradient loss alone.
However, we observed that including the L1 reconstruction term tends to improve the convergence speed consistently across all models.
In the next experiments \textit{(f,g)}, we replace SEA-RAFT \cite{Wang:2024:SEA} with the RAFT model \cite{TeedD20} and the unsupervised SMURF \cite{Stone:2021:SMURF}, respectively. 
The drop in the VOS accuracy is not significant, which demonstrates that obtaining \oursName is not sensitive to a specific choice of the flow model.
\cref{fig:flow_approx_err} further examines the training resilience to inaccurate target flow.
In both examples, the optical flow from the pre-trained network is inaccurate (even artefact-prone), yet it has no visible effect on the quality of \oursName.
Curiously, the second example in \cref{fig:flow_approx_err} also reveals one limitation of \oursName: the apparent motion of the tail and the head of the snake have opposite directions, which decouples their feature representation. 

Finally, \textit{(h,i)} we test two strategies  for sampling frame pairs from a video in \cref{tab:ablation}.
Our base configuration uses a temporal window of 5 frames and can select frame $t^\prime$ either from the past or the future.
We increase this temporal window to 9 \textit{(h)}, which leads to a slight deterioration of VOS accuracy -- presumably due to the more challenging estimation of optical flow.
The setting \textit{(i)} selects the immediate next future frame as $t^\prime$.
Here, the VOS accuracy barely changes.
This implies that \textit{(i)} \oursName does not simply overfit apparent motion (compare to \textit{(a)}), and \textit{(ii)} the motion samples \emph{across the dataset}, not just a temporal window, play a more critical role in embedding motion profiles.

\inparagraph{Further study.}
Our further analysis, provided in the supplemental material, shows that:
\textit{(1)} \oursName scales well with the input resolution, further improving the VOS accuracy when the input resolution is doubled (\cf \cref{supp-tab:ablation_knn});
\textit{(2)} the accuracy gains from \oursName do not arise merely from higher resolution of the feature map per se, but from its complementary motion-derived properties (\cf \cref{supp-tab:ablation_lin_down});
\textit{(3)} \oursName also scales effectively to larger transformer models (\eg ViT-L) (\cf \cref{tab:large_backbones}).

\section{Limitations}
\label{sec:limitations}

\inparagraph{Application scope.}
\oursName relies on a pre‑trained optical flow network and video data for training.
It assumes either brightness constancy in the video stream or availability of synthetic data for pre-training the optical flow model.
While these assumptions generally hold for standard RGB videos, they may not apply in other domains, such as medical imaging (\eg MRI, CT), thermal imaging or low-light scenarios. 

\begin{wraptable}{r}{0.6\textwidth}
\centering
\vspace{-1em}
\scriptsize
\begin{tabular}{lccccc}
\toprule
Model & Person & Wall & Landscape & Vegetation & Ground \\
\midrule
DINO‑S16 & 69.3 & 46.5 & 43.9 & 65.5 & 33.3 \\
\rowcolor{azure}
\quad + \oursName\text{-YT} & \textbf{75.6} & \textbf{50.0} & \textbf{50.8} & \textbf{69.9} & \textbf{37.0} \\
\midrule
DINO‑B16 & 72.9 & 51.4 & 51.0 & 70.3 & 38.9 \\
\rowcolor{azure}
\quad + \oursName\text{-K} & \textbf{77.8} & \textbf{52.9} & \textbf{53.1} & \textbf{71.7} & \textbf{39.6} \\
\midrule
DINOv2‑S14 & 76.9 & 57.6 & 59.2 & 71.0 & 44.6 \\
\rowcolor{azure}
\quad + \oursName\text{-YT} & \textbf{81.7} & \textbf{59.3} & \textbf{60.5} & \textbf{73.0} & \textbf{45.7} \\
\midrule
DINOv2‑B14 & 77.0 & 59.2 & 59.6 & 70.3 & 44.9 \\
\rowcolor{azure}
\quad + \oursName\text{-K} & \textbf{83.0} & \textbf{61.7} & \textbf{61.3} & \textbf{72.3} & \textbf{45.1} \\
\midrule
MAE‑B16 & 72.2 & 50.8 & 52.7 & 66.9 & 36.1 \\
\rowcolor{azure}
\quad + \oursName\text{-K} & \textbf{78.6} & \textbf{51.3} & \textbf{53.7} & \textbf{69.9} & \textbf{38.8} \\
\bottomrule
\end{tabular}
\caption{\textbf{Semantic segmentation accuracy on COCO-Stuff (IoU, \%).}
As expected from motion parallax, the gains on (potentially) dynamic classes (\eg ``person'') are larger compared to that of typical background categories (\eg ``vegetation''). Nevertheless, \oursName leads to a consistent segmentation improvement across \emph{all} categories.}
\label{tab:motion_bias}
\end{wraptable}


\inparagraph{Frozen backbone.}
Recall that training \oursName involves updating only the decoder parameters, while keeping the encoder parameters fixed. 
Consequently, the encoder representation imposes an upper bound on \oursName's downstream accuracy, especially in terms of high‑frequency content. 
Although we have shown that \oursName generalises across widely used self-supervised encoders, such as MAE~\cite{He:2022:MAA}, DINO~\cite{Caron:2021:EPS}, DINOv2~\cite{Oquab:2023:DINOv2} and across different model capacities, \oursName may be less effective with backbones that underrepresent high-frequency details in their intermediate feature maps.

\inparagraph{Motion bias.}
Owing to its training approach, \oursName tends to emphasise image regions with larger magnitudes of expected motion,  typically corresponding to foreground dynamic objects, relative to the static background areas.
To quantitatively assess this behaviour, we report per-category IoU scores on COCO-Stuff in \cref{tab:motion_bias}, following the probing protocol described in \cref{sec:experiments}. 
We observe that the improvement on the ``person'' category is indeed more pronounced than for static classes.
Nevertheless, \oursName yields consistent accuracy gains across all categories, regardless of whether they are static or dynamic in nature.

%% file: sections/05_conclusion.tex
\section{Conclusion}

We presented \oursName, a pixel-dense and versatile representation embedding motion profiles.
Our experiments provide compelling evidence that \oursName enhances the representation power of pre-trained encoders across all downstream tasks considered in our study.
Specifically, \oursName possesses temporal consistency and exhibits a remarkable level of spatial detail, encompassing semantic and geometric cues without explicit supervision.
More broadly, our work addresses motion stochasticity in a principled fashion, revealing a powerful synergy between optical flow networks and large video datasets.
\oursName takes a significant step towards label-efficient and versatile models for high-precision tasks, such as image-based 3D reconstruction, object-level segmentation and tracking.

{\footnotesize
\inparagraph{Acknowledgements.} This work was supported by the ERC Advanced Grant SIMULACRON and DFG project CR 250/26-1 ``4D-YouTube''. NA thanks Junhwa Hur and Jochen Gast for their valuable feedback.
}

%% file: supp-arxiv.tex
\section{Qualitative Examples}
\label{supp-sec:videos}

Our \href{https://cvg.cit.tum.de/webshare/g/papers/flowfeat/supp_videos.zip}{supplemental material (.zip, 90MB)}  is accompanied by a qualitative video comparison.
We select multiple validation sequences from DAVIS-2017 \cite{PontTuset2017} and run linear probing  using DINO2-S14 \cite{Oquab:2023:DINOv2} as the baseline encoder.
\begin{itemize}
    \item \verb|a_vos_featup.mp4| compares \oursName against FeatUp \cite{Fu:2024:FeatUp}. The videos also visualise the ground-truth segmentation masks and the first three principal components of \oursName representation. \oursName achieves compelling segmentation accuracy, remains sharp at object boundaries and more temporally stable than FeatUp.
    \item \verb|b_pca_featup.mp4| inspects the first three principal components of \oursName and FeatUp for two input resolutions: $224 \times 400$ and $480 \times 854$. We observe that \oursName scales gracefully and reveals a greater level of detail. By contrast, FeatUp struggles to adapt to the higher resolution and exhibits static artefacts. Note that both \oursName and FeatUp were trained on $224 \times 224$ image crops. 
    \item \verb|c_pca_video_models.mp4| compares \oursName to VideoMAE \cite{tong:2022:videomae} based on ViT-B16 and V-JEPA \cite{Bardes:2024:RFP} based on ViT-L16, which are pre-trained on videos in a self-supervised manner. We observe that neither of these two models is capable of producing a satisfying level of granularity in the feature maps. Additionally, V-JEPA exhibits peculiar artefacts, which make it unsuitable for downstream dense tasks. In contrast, \oursName leverages video datasets more effectively, producing fine-grained and temporally stable feature representations.
    \item Finally, \verb|d_pca_vs_loftup.mp4| compares \oursName to a concurrent work, LoftUp \cite{Haiwen:2025:LUP}, \emph{pre-trained with mask supervision} via the Segment Anything model \cite{kirillov:2023:SAM}. LoftUp offers an improved representation over FeatUp and exhibits a high discrimination level of the background regions. However, LoftUp suffers from artefacts, often producing ragged-looking boundaries of moving objects and background elements. Without relying on mask annotation, \oursName demonstrates consistent spatial and temporal precision when it comes to dynamic objects.  For example, observe sharper boundaries in the ``goat'' sequence, and a more distinguishable representation of the car and the dancers. These observations explain the quantitative advantage of \oursName over LoftUp on the VOS benchmark (\cf \cref{tab:vos_results}).
\end{itemize}

\section{Further Implementation Details and Analyses}
\label{supp-sec:impl_details}

\subsection{Video object segmentation}
\label{supp-sec:impl_details_vos}

\inparagraph{Linear Probing.} Linear probing maps the learned representations to object masks. The linear layer has $(d + 1) \times (C + 1)$ parameters, accounting for the bias term and the background class.
The training process of the linear probe is standardised across all evaluated methods.
Using the cross-entropy loss, we employ the Adam solver \cite{Kingma:2015:AAM} and train the linear projection for $500$ iterations with a learning rate of $5\times10^{-3}$ and a weight decay of $5\times10^{-4}$.
We run inference and training of the linear probe by fixing the image height at $480p$ and adjusting the width to be divisible by 64, as done by \citet{Caron:2021:EPS}. We apply linear probing on the native feature resolution produced by the model.
If necessary, we resize the predicted masks to the original image resolution to compute the metrics \wrt the original ground truth masks.

Both \oursName and FeatUp \cite{Fu:2024:FeatUp} utilise a two-stage architecture: a pre-trained encoder that produces low-resolution features, followed by a decoder that generates high-resolution representations.
To evaluate the contribution of the high-resolution features, we combine them with the encoder's low-resolution feature map.
In detail, we bilinearly upsample the backbone features and concatenate them with the high-resolution feature tensors.
The linear layer projects this joint feature representation to the segmentation logits for each pixel.

Note that linear probing is not auto-regressive -- in contrast to local KNN.
Specifically, we run inference with the same pre-trained linear projection on all remaining frames in the video.
Intuitively, linear probing in the context of VOS is akin to few-shot learning, and provides a more interpretable measure of the spatio-temporal representation quality. 

To assess whether the improved VOS performance of \oursName is due to the higher feature resolution alone, we downsample the output features of \oursName and FeatUp to the same resolution as the baseline encoder (DINO2-S14~\cite{Oquab:2023:DINOv2}). This enables us to directly compare the models under the identical conditions of the feature resolution.
\cref{supp-tab:ablation_lin_down} reports the results.
While all methods experience some drop in performance -- as expected -- both \oursName variants significantly outperform the baseline and FeatUp. This result demonstrates that \oursName yields accuracy gains not merely due to the higher resolution, but also due to the complementary properties derived from motion, which the encoder's representation lacks.

\begin{table}[t]
\centering
\caption{\textbf{(\subref{supp-tab:ablation_lin_down}) Effect of reduced feature resolution in linear probing (DAVIS-2017, val).} 
All features are downsampled to match the baseline encoder resolution. 
$\Delta$ indicates the absolute change in the VOS accuracy \wrt the original full-resolution setting of each model. 
Despite the lower resolution, \oursName retains strong VOS performance, confirming that high resolution alone does not explain the observed benefits; \oursName embeds a motion-based feature modality that is complementary to the encoder's representation.
\textbf{(\subref{tab:large_backbones}) \oursName generalizes to larger backbones (DAVIS-2017, val).} 
As expected, training \oursName with ViT-L leads to a consistent improvement of VOS accuracy over the baseline -- both for MAE and DINOv2.
}
  \label{tab:extra_ablation}
  
\begin{subtable}[t]{0.59\linewidth}
\caption{}
\label{supp-tab:ablation_lin_down}
\newcolumntype{Y}{S[table-format=2.1]@{\hspace{0.4em}}} 
\newcolumntype{D}{S[table-format=+1.1]@{\hspace{1.3em}}} 
\newcolumntype{Z}{S[table-format=+1.1]} 
\setlength{\tabcolsep}{3pt}
\medskip
\footnotesize
\centering
\begin{tabularx}{\linewidth}{lXYZ@{\hspace{0.8em}}YZ@{\hspace{0.8em}}YZ}
\toprule
{Method} & {Scale} & $\mathcal{JF}_m$ & $/\Delta$ & $\mathcal{J}_m$ & $/\Delta$ & $\mathcal{F}_m$ & $/\Delta$ \\
\midrule
DINO2-S14 \cite{Oquab:2023:DINOv2} &  {--} &   57.5  & {--} &  54.2 & {--} & 60.7 & {--} \\
\leftPad FeatUp~\cite{Fu:2024:FeatUp}   &  $\downarrow$x14 & 59.5 & \minus{1.0} & \bfseries 56.4 & \minus{1.0} & 62.6 & \minus{1.0} \\ 
\rowcolor{azure}
\leftPad\textbf{\oursName-YT}               &  $\downarrow$x14 & 62.2 & \minus{3.6} & \bfseries 58.2 & \minus{3.8} &  66.3 & \minus{3.4} \\ 
\rowcolor{azure}
\leftPad\textbf{\oursName-K}                &  $\downarrow$x14 & \bfseries 62.3 & \minus{2.3} & 58.1 & \minus{2.9} & \bfseries 66.4 & \minus{1.8} \\ 
\bottomrule
\end{tabularx}
\end{subtable}
  \hfill
\begin{subtable}[t]{0.39\linewidth}
\caption{}
\label{tab:large_backbones}
\setlength{\tabcolsep}{3pt}
\newcolumntype{Y}{S[table-format=2.1]}
\footnotesize
\centering
\begin{tabularx}{\linewidth}{lYYY}
\toprule
{Method} & {$\mathcal{JF}_m$} & {$\mathcal{J}_m$} & {$\mathcal{F}_m$} \\
\midrule
DINOv2-L14~\cite{Oquab:2023:DINOv2} & 59.4 & 55.8 & 63.0 \\
\rowcolor{azure}
\leftPad \bfseries \oursName-YT & \bfseries 66.9 & \bfseries 63.4 & \bfseries 70.4 \\
\midrule
MAE-L14~\cite{He:2022:MAA} & 46.7 & 44.4 & 49.0 \\
\rowcolor{azure}
\leftPad \bfseries \oursName-YT & \bfseries 55.4 & \bfseries 52.0 & \bfseries 58.9 \\
\bottomrule
\end{tabularx}
\end{subtable}
\vspace{-0.8em}
\end{table}

To evaluate generalizion of \oursName to the backbone architectures with larger capacity, we consider ViT-L and train \oursName by boostrapping it from DINO2-L14~\cite{Oquab:2023:DINOv2} and MAE-L14~\cite{He:2022:MAA}. We evaluate the models using the same linear probing protocol on VOS.
As shown in \cref{tab:large_backbones}, \oursName leads to substantial performance gains for both models.
These results confirm that \oursName is effective across different encoder architectures, with varying model capacities and pre-training schemes.

\inparagraph{Local KNN.} We adopt the label propagation approach from \citet{Caron:2021:EPS}. This protocol requires downsampling the high-resolution feature maps to match the encoder feature resolution.
We apply label propagation independently on both the backbone and the downsampled features, and compute the mean of the resulting logits. This ensures hyperparameter consistency of our local KNN evaluation with previous work and across model architectures.

\begin{table}[t]
\caption{\textbf{Scaling up the feature resolution in local KNN probing (DAVIS-2017, val).} We increase the feature resolution by a factor of two and re-run the local KNN unchanged otherwise. The $\Delta$ reports the absolute difference in the corresponding metric \wrt the base setting of local KNN. CRW~\cite{Jabri:2020:STC} is a CNN-based approach provided for a reference. The encoder and FeatUp do not benefit from the higher feature resolution. By contrast, \oursName considerably improves its VOS accuracy.}
\label{supp-tab:ablation_knn}

\newcolumntype{Y}{S[table-format=2.1]@{\hspace{0.4em}}} 
\newcolumntype{D}{S[table-format=+1.1]@{\hspace{1.3em}}} 
\newcolumntype{Z}{S[table-format=+1.1]} 
\setlength{\tabcolsep}{3pt}
\medskip
\footnotesize
\centering
\begin{tabularx}{\linewidth}{lXYZ@{\hspace{0.8em}}YZ@{\hspace{0.8em}}YZ}
\toprule
{Method} & {Scale} & $\mathcal{JF}_m$ & $/\Delta$ & $\mathcal{J}_m$ & $/\Delta$ & $\mathcal{F}_m$ & $/\Delta$ \\
\midrule
CRW-Res18~\cite{Jabri:2020:STC} & $\times 2$ & 65.2 & & 63.1 &  & 67.3 & \\
\midrule
DINO2-S14 \cite{Oquab:2023:DINOv2} &  $\times 2$ &    63.3 & \minus{1.8} & 62.8 & \minus{0.9} & 63.7 & \minus{2.9} \\
\leftPad FeatUp~\cite{Fu:2024:FeatUp}   &  $\times 2$ & 64.6 & \minus{0.9} & \bfseries 64.5 & \minus{0.5} & 64.6 & \minus{1.5} \\ 
\rowcolor{azure}
\leftPad\textbf{\oursName-YT}               &  $\times 2$ & \bfseries 70.3 & \plus{2.7} & 68.0 & \plus{2.4} & \bfseries 72.5 & \plus{2.9} \\ 
\rowcolor{azure}
\leftPad\textbf{\oursName-K}                &  $\times 2$ & \bfseries 70.0 & \plus{1.5} & 67.5 & \plus{1.4} & \bfseries 72.5 & \plus{1.6} \\ 
\bottomrule
\end{tabularx}
\end{table}

Since downsampling the high-resolution features goes against our motivation, we analyse the impact of the increased feature resolution in \cref{supp-tab:ablation_knn}.
Here, we keep the hyperparameters of the local KNN from the base setting above, but increase the resolution of the feature maps by a factor of 2.
As a reference, we provide the VOS accuracy of CRW \cite{Jabri:2020:STC}, a CNN-based approach producing the feature maps at the required resolution. 
Surprisingly, neither the encoder nor FeatUp benefit from the resolution increase. 
By contrast, \oursName improves the results further by significant margins and substantially outpeforms the CNN-based reference.


\subsection{Semantic segmentation}
\label{supp-sec:impl_details_seg}

We keep the DPT decoder in \oursName frozen and only train a shallow, one-layer probe.
For the low-resolution encoder features and the high-dimensional FeatUp representation we use linear probing.
For \oursName, we complement the predictions from the encoder with the predictions provided by attention probing.
Our implementation of attention probing is inspired by previous work \cite{Bardes:2024:RFP}.
In our evaluation, we extend this technique to dense prediction tasks.
We initialise $C=27$ (the number of semantic categories) learnable queries in the probe.
Each query has the dimensionality of $d$, matching that of \oursName.
We use a single block of cross-attention to condition the queries on the \oursName representation.
Finally, we compute the dot product of the conditioned queries with the spatial feature grid produced by \oursName.
As a result, we obtain a prediction of size $C \times H \times W$.
We also found that downsampling the \oursName maps in the cross-attention block significantly improves the probe efficiency without detriment to the probing accuracy.

We train the models with Adam \cite{Kingma:2015:AAM} using the cross-entropy loss.
We sample mini-batches of size $32$, setting the learning rate to $10^{-4}$ and weight decay to $0.001$.
All models tend to converge within 100K iterations, which takes less than a day on a single GPU -- except for FeatUp, which runs longer, as we discuss in the next section.

For the post-hoc refinement (the ``++'' variants of \oursName), we use a simplified PAMR implementation \cite{Araslanov:2020:SSS}.
Specifically, we use a single kernel of size $11 \times 11$ and a fixed scaling factor of $0.1$ to produce the local affinity distribution.
Crucially, we do not leverage the image intensities to compute the distribution, but replace it with the \oursName representation.
The refinement runs for $10$ steps.

\subsection{Monocular depth estimation}
\label{supp-sec:impl_details_depth}

The experiments on monocular depth largely follow the setting of the semantic segmentation above.
The only conceptual difference is the definition of $C$, which in monocular depth stands for the number of depth bins. 
Specifically, to predict the AdaBins \cite{Bhat2020AdaBinsDE} representation with $C = 256$ bins (\ie a tensor of size $C \times H \times W$), we initialise $C = 256$ learnable queries in the probe and pass them through a single block of cross-attention.
As in semantic segmentation, we compute the dot product of the conditioned queries with the \oursName tensor.
We train the probes with Adam optimizer \cite{Kingma:2015:AAM}, batch size $16$ and set the learning rate to $10^{-4}$ and weight decay to $10^{-5}$.
Following previous work \cite[][A.3.1]{Banani:2024:P3D}, we use a combination of the scale-invariant depth loss \cite{Eigen:2014:DMP} and the gradient matching loss \cite{Li:2018:MDL}.
We train all models for up to 100K iterations.

\section{Efficiency Analysis}
\label{supp-sec:efficiency}

 We analyse the computational and runtime efficiency of \oursName and compare those to the baseline encoder DINO2-S14~\cite{Oquab:2023:DINOv2} and FeatUp \cite{Fu:2024:FeatUp}.
 Specifically, we set the input resolution to $224 \times 224$ and measure floating-point operations (FLOPs) as well as the FPS rate on an RTX 8000 GPU with $48$GB of memory.
 \cref{tab:complexity} summarises the benchmark results.
In fact, \oursName incurs more FLOPs in total than FeatUp.
However, all operations in the DPT decoder of \oursName are highly parallelisable, hence efficient. As a result, \oursName achieves a significantly higher throughput.
In the practical terms of FPS, \oursName runs four times faster than FeatUp. Indeed, FeatUp’s implementation of the bilateral upsampler, though impressively more efficient than previous work, still falls short of the DPT runtime efficiency.

\begin{table}[t]
\caption{\textbf{Computational and runtime complexity.} We compare the total and decoder-only floating-point operations (FLOPs), as well as the throughput measured by the frames per second (FPS) rate. Here, we use the DINO2-S14~\cite{Oquab:2023:DINOv2} baseline, the input resolution of $224 \times 224$ and RTX 8000 GPU.}
\label{tab:complexity}

\setlength{\tabcolsep}{5pt}
\newcolumntype{Y}{S[table-format=3.2]} 
\medskip
\footnotesize
\centering
\begin{tabularx}{\linewidth}{XcccY}
\toprule
Method & {Total FLOPs} & {Decoder FLOPs}  & {FPS} \\
\midrule
DINO2-S14~\cite{Oquab:2023:DINOv2} & 6.14B & {--}  & 176.79 \\
\leftPad  FeatUp~\cite{Fu:2024:FeatUp} & 16.54B & 10.33B & 25.12 \\
\rowcolor{azure}
\leftPad \oursName & 23.43B & 17.3B & 105.82 \\
\bottomrule
\end{tabularx}
\end{table}